\documentclass[10pt,twocolumn,letterpaper]{article}
\usepackage[T1]{fontenc}
\usepackage[applications]{wacv}
\usepackage{float}
\usepackage{placeins}
\usepackage{tabularx}
\definecolor{wacvblue}{rgb}{0.21,0.49,0.74}
\usepackage[pagebackref,breaklinks,colorlinks,allcolors=wacvblue]{hyperref}
\backrefparscanfalse
\AddToHook{para/begin}{\looseness=-1}

\def\wacvPaperID{2337}
\def\confName{WACV}
\def\confYear{2027}

\newcommand{\RegTireMean}{0.697}
\newcommand{\RegTireStd}{0.045}

\newcommand{\RegInteriorMean}{0.613}
\newcommand{\RegInteriorStd}{0.032}

\newcommand{\RegFoodMean}{0.705}
\newcommand{\RegFoodStd}{0.009}

\newcommand{\RegSoapMean}{0.909}
\newcommand{\RegSoapStd}{0.006}

\newcommand{\ControlMAEMax}{0.013}
\newcommand{\ControlRhoAll}{0.980}
\newcommand{\ControlCells}{200}
\newcommand{\ReachTolerance}{0.05}
\newcommand{\ControlRhoCeiling}{0.980}
\newcommand{\ControlWorstCell}{0.0331}
\newcommand{\ControlWorstAttribute}{food}
\newcommand{\ControlMedianMax}{0.0124}
\newcommand{\ControlPNinetyMax}{0.0224}
\newcommand{\ExtTirePromptMAE}{0.2672}
\newcommand{\ExtTirePromptMAESD}{0.0303}
\newcommand{\ExtTirePromptRho}{-0.061}

\newcommand{\ExtTireMPGDMAE}{0.0031}
\newcommand{\ExtTireMPGDMAESD}{0.0015}
\newcommand{\ExtTireMPGDRho}{0.980}

\newcommand{\ExtTireUGMAE}{0.1788}
\newcommand{\ExtTireUGMAESD}{0.0279}
\newcommand{\ExtTireUGRho}{0.764}

\newcommand{\ExtTireOursMAE}{0.0129}
\newcommand{\ExtTireOursMAESD}{0.0028}
\newcommand{\ExtTireOursRho}{0.980}
\newcommand{\ExtTireOursN}{50}

\newcommand{\ExtInteriorPromptMAE}{0.3905}
\newcommand{\ExtInteriorPromptMAESD}{0.0592}
\newcommand{\ExtInteriorPromptRho}{-0.039}

\newcommand{\ExtInteriorMPGDMAE}{0.0242}
\newcommand{\ExtInteriorMPGDMAESD}{0.0384}
\newcommand{\ExtInteriorMPGDRho}{0.961}

\newcommand{\ExtInteriorUGMAE}{0.1791}
\newcommand{\ExtInteriorUGMAESD}{0.0551}
\newcommand{\ExtInteriorUGRho}{0.789}

\newcommand{\ExtInteriorOursMAE}{0.0069}
\newcommand{\ExtInteriorOursMAESD}{0.0009}
\newcommand{\ExtInteriorOursRho}{0.980}

\newcommand{\ExtFoodPromptMAE}{0.2600}
\newcommand{\ExtFoodPromptMAESD}{0.0187}
\newcommand{\ExtFoodPromptRho}{-0.073}

\newcommand{\ExtFoodMPGDMAE}{0.0149}
\newcommand{\ExtFoodMPGDMAESD}{0.0118}
\newcommand{\ExtFoodMPGDRho}{0.980}

\newcommand{\ExtFoodUGMAE}{0.1866}
\newcommand{\ExtFoodUGMAESD}{0.0354}
\newcommand{\ExtFoodUGRho}{0.664}

\newcommand{\ExtFoodOursMAE}{0.0086}
\newcommand{\ExtFoodOursMAESD}{0.0026}
\newcommand{\ExtFoodOursRho}{0.980}

\newcommand{\ExtSoapPromptMAE}{0.2600}
\newcommand{\ExtSoapPromptMAESD}{0.0216}
\newcommand{\ExtSoapPromptRho}{0.027}

\newcommand{\ExtSoapMPGDMAE}{0.0134}
\newcommand{\ExtSoapMPGDMAESD}{0.0114}
\newcommand{\ExtSoapMPGDRho}{0.970}

\newcommand{\ExtSoapUGMAE}{0.1816}
\newcommand{\ExtSoapUGMAESD}{0.0396}
\newcommand{\ExtSoapUGRho}{0.718}

\newcommand{\ExtSoapOursMAE}{0.0072}
\newcommand{\ExtSoapOursMAESD}{0.0020}
\newcommand{\ExtSoapOursRho}{0.980}

\newcommand{\AblationCondOnlyMAE}{0.1960}
\newcommand{\AblationCondOnlyMAESD}{0.0482}

\newcommand{\AblationRawMAE}{0.1789}
\newcommand{\AblationRawMAESD}{0.0316}

\newcommand{\AblationConstantMAE}{0.0715}
\newcommand{\AblationConstantMAESD}{0.0110}

\newcommand{\AblationAudFull}{0.285}
\newcommand{\AblationAudFullSD}{0.209}
\newcommand{\AblationAudGOnly}{0.041}
\newcommand{\AblationAudGOnlySD}{0.075}
\newcommand{\AblationAudCondOnly}{0.184}
\newcommand{\AblationAudCondOnlySD}{0.181}
\newcommand{\AblationAudRaw}{0.254}
\newcommand{\AblationAudRawSD}{0.202}
\newcommand{\AblationAudConstant}{0.159}
\newcommand{\AblationAudConstantSD}{0.127}
\newcommand{\AblationGridFullMAE}{0.0129}
\newcommand{\AblationGridFullMAESD}{0.0028}
\newcommand{\AblationGridGOnlyMAE}{0.0154}
\newcommand{\AblationGridGOnlyMAESD}{0.0041}
\newcommand{\RoundTripTirePct}{0.91}
\newcommand{\RoundTripInteriorPct}{1.83}
\newcommand{\RoundTripFoodPct}{1.16}
\newcommand{\RoundTripSoapPct}{0.42}

\newcommand{\RoundTripPctMax}{1.84}
\newcommand{\RoundTripRMin}{0.9986}
\newcommand{\RoundTripShuffledMax}{6\%}
\newcommand{\RoundTripImages}{847}
\newcommand{\AudTirePrompt}{-0.014}
\newcommand{\AudTirePromptSD}{0.081}
\newcommand{\AudTireMPGD}{-0.021}
\newcommand{\AudTireMPGDSD}{0.095}
\newcommand{\AudTireUG}{0.096}
\newcommand{\AudTireUGSD}{0.196}
\newcommand{\AudTireOurs}{0.285}
\newcommand{\AudTireOursSD}{0.209}
\newcommand{\AudInteriorPrompt}{0.001}
\newcommand{\AudInteriorPromptSD}{0.056}
\newcommand{\AudInteriorMPGD}{0.055}
\newcommand{\AudInteriorMPGDSD}{0.062}
\newcommand{\AudInteriorUG}{0.185}
\newcommand{\AudInteriorUGSD}{0.171}
\newcommand{\AudInteriorOurs}{0.410}
\newcommand{\AudInteriorOursSD}{0.082}
\newcommand{\AudFoodPrompt}{-0.017}
\newcommand{\AudFoodPromptSD}{0.083}
\newcommand{\AudFoodMPGD}{0.124}
\newcommand{\AudFoodMPGDSD}{0.074}
\newcommand{\AudFoodUG}{0.076}
\newcommand{\AudFoodUGSD}{0.106}
\newcommand{\AudFoodOurs}{0.187}
\newcommand{\AudFoodOursSD}{0.100}
\newcommand{\AudSoapPrompt}{0.027}
\newcommand{\AudSoapPromptSD}{0.041}
\newcommand{\AudSoapMPGD}{0.078}
\newcommand{\AudSoapMPGDSD}{0.081}
\newcommand{\AudSoapUG}{0.312}
\newcommand{\AudSoapUGSD}{0.197}
\newcommand{\AudSoapOurs}{0.452}
\newcommand{\AudSoapOursSD}{0.137}
\newcommand{\AudSeeds}{10}
\newcommand{\AudBackbone}{SigLIP}
\newcommand{\AudRtwoMin}{0.61}
\newcommand{\HumanParticipants}{15}
\newcommand{\HumanImages}{40}
\newcommand{\HumanPresentations}{48}
\newcommand{\HumanRepeats}{8}

\newcommand{\HumanTireSpearman}{0.827}
\newcommand{\HumanTireLinearRtwo}{0.665}

\newcommand{\HumanSoapSpearman}{0.860}
\newcommand{\HumanSoapLinearRtwo}{0.786}

\newcommand{\HumanTirePoints}{5}
\newcommand{\HumanSoapPoints}{7}

\newcommand{\MPGDStrongTire}{300}
\newcommand{\ExtTireMPGDsMAE}{0.1276}
\newcommand{\ExtTireMPGDsMAESD}{0.0906}
\newcommand{\ExtTireMPGDsRho}{0.862}
\newcommand{\AudTireMPGDs}{0.053}
\newcommand{\AudTireMPGDsSD}{0.139}

\newcommand{\MPGDStrongInterior}{100}
\newcommand{\ExtInteriorMPGDsMAE}{0.1304}
\newcommand{\ExtInteriorMPGDsMAESD}{0.0655}
\newcommand{\ExtInteriorMPGDsRho}{0.869}
\newcommand{\AudInteriorMPGDs}{0.157}
\newcommand{\AudInteriorMPGDsSD}{0.102}

\newcommand{\MPGDStrongFood}{100}
\newcommand{\ExtFoodMPGDsMAE}{0.0832}
\newcommand{\ExtFoodMPGDsMAESD}{0.0470}
\newcommand{\ExtFoodMPGDsRho}{0.906}
\newcommand{\AudFoodMPGDs}{0.134}
\newcommand{\AudFoodMPGDsSD}{0.118}

\newcommand{\MPGDStrongSoap}{30}
\newcommand{\ExtSoapMPGDsMAE}{0.0702}
\newcommand{\ExtSoapMPGDsMAESD}{0.0256}
\newcommand{\ExtSoapMPGDsRho}{0.936}
\newcommand{\AudSoapMPGDs}{0.152}
\newcommand{\AudSoapMPGDsSD}{0.103}
\newcommand{\AblInteriorGOnlyMAE}{0.0084}
\newcommand{\AblInteriorGOnlyMAESD}{0.0023}
\newcommand{\AblInteriorGOnlyAud}{0.160}
\newcommand{\AblInteriorGOnlyAudSD}{0.120}
\newcommand{\AblInteriorCondOnlyMAE}{0.2011}
\newcommand{\AblInteriorCondOnlyMAESD}{0.0496}
\newcommand{\AblInteriorCondOnlyAud}{0.295}
\newcommand{\AblInteriorCondOnlyAudSD}{0.174}
\newcommand{\AblInteriorRawMAE}{0.1801}
\newcommand{\AblInteriorRawMAESD}{0.0463}
\newcommand{\AblInteriorRawAud}{0.320}
\newcommand{\AblInteriorRawAudSD}{0.113}
\newcommand{\AblInteriorConstantMAE}{0.0665}
\newcommand{\AblInteriorConstantMAESD}{0.0105}
\newcommand{\AblInteriorConstantAud}{0.060}
\newcommand{\AblInteriorConstantAudSD}{0.065}
\newcommand{\AblFoodGOnlyMAE}{0.0084}
\newcommand{\AblFoodGOnlyMAESD}{0.0041}
\newcommand{\AblFoodGOnlyAud}{0.090}
\newcommand{\AblFoodGOnlyAudSD}{0.095}
\newcommand{\AblFoodCondOnlyMAE}{0.2406}
\newcommand{\AblFoodCondOnlyMAESD}{0.0647}
\newcommand{\AblFoodCondOnlyAud}{0.079}
\newcommand{\AblFoodCondOnlyAudSD}{0.139}
\newcommand{\AblFoodRawMAE}{0.2146}
\newcommand{\AblFoodRawMAESD}{0.0490}
\newcommand{\AblFoodRawAud}{0.084}
\newcommand{\AblFoodRawAudSD}{0.115}
\newcommand{\AblFoodConstantMAE}{0.0769}
\newcommand{\AblFoodConstantMAESD}{0.0199}
\newcommand{\AblFoodConstantAud}{0.004}
\newcommand{\AblFoodConstantAudSD}{0.090}
\newcommand{\AblSoapGOnlyMAE}{0.0086}
\newcommand{\AblSoapGOnlyMAESD}{0.0023}
\newcommand{\AblSoapGOnlyAud}{0.114}
\newcommand{\AblSoapGOnlyAudSD}{0.088}
\newcommand{\AblSoapCondOnlyMAE}{0.1594}
\newcommand{\AblSoapCondOnlyMAESD}{0.0616}
\newcommand{\AblSoapCondOnlyAud}{0.465}
\newcommand{\AblSoapCondOnlyAudSD}{0.169}
\newcommand{\AblSoapRawMAE}{0.1780}
\newcommand{\AblSoapRawMAESD}{0.0583}
\newcommand{\AblSoapRawAud}{0.367}
\newcommand{\AblSoapRawAudSD}{0.146}
\newcommand{\AblSoapConstantMAE}{0.0825}
\newcommand{\AblSoapConstantMAESD}{0.0197}
\newcommand{\AblSoapConstantAud}{0.175}
\newcommand{\AblSoapConstantAudSD}{0.104}
\newcommand{\AblAllFullMAELo}{0.007}
\newcommand{\AblAllFullMAEHi}{0.013}
\newcommand{\AblAllGOnlyMAELo}{0.008}
\newcommand{\AblAllGOnlyMAEHi}{0.015}
\newcommand{\AblAllCondOnlyMAELo}{0.159}
\newcommand{\AblAllCondOnlyMAEHi}{0.241}
\newcommand{\AblAllRawMAELo}{0.178}
\newcommand{\AblAllRawMAEHi}{0.215}
\newcommand{\AblAllConstantMAELo}{0.067}
\newcommand{\AblAllConstantMAEHi}{0.082}
\newcommand{\AblAllCondOnlyAudLo}{0.079}
\newcommand{\AblAllCondOnlyAudHi}{0.465}

\IfFileExists{twoafc_metrics.tex}{\newcommand{\TwoAFCTirePrompt}{0.98}
\newcommand{\TwoAFCTirePromptLo}{0.94}
\newcommand{\TwoAFCTirePromptHi}{1.00}
\newcommand{\TwoAFCTireMPGD}{0.95}
\newcommand{\TwoAFCTireMPGDLo}{0.90}
\newcommand{\TwoAFCTireMPGDHi}{0.99}
\newcommand{\TwoAFCTireMPGDs}{0.98}
\newcommand{\TwoAFCTireMPGDsLo}{0.92}
\newcommand{\TwoAFCTireMPGDsHi}{1.00}
\newcommand{\TwoAFCTireUG}{0.85}
\newcommand{\TwoAFCTireUGLo}{0.71}
\newcommand{\TwoAFCTireUGHi}{0.95}
\newcommand{\TwoAFCTireRaters}{15}

\newcommand{\TwoAFCSoapPrompt}{0.98}
\newcommand{\TwoAFCSoapPromptLo}{0.93}
\newcommand{\TwoAFCSoapPromptHi}{1.00}
\newcommand{\TwoAFCSoapMPGD}{0.97}
\newcommand{\TwoAFCSoapMPGDLo}{0.92}
\newcommand{\TwoAFCSoapMPGDHi}{1.00}
\newcommand{\TwoAFCSoapMPGDs}{0.97}
\newcommand{\TwoAFCSoapMPGDsLo}{0.90}
\newcommand{\TwoAFCSoapMPGDsHi}{1.00}
\newcommand{\TwoAFCSoapUG}{0.86}
\newcommand{\TwoAFCSoapUGLo}{0.77}
\newcommand{\TwoAFCSoapUGHi}{0.94}
\newcommand{\TwoAFCSoapRaters}{15}

\newcommand{\TwoAFCMinPct}{84}
\newcommand{\TwoAFCMaxPct}{98}
}{\input{twoafc_pending.tex}}

\title{ScoreDial: Precise Control of Visual Attributes in Image Generation\\
via Regressor Guidance from Limited Human Ratings}

\author{%
Jianbo Tang$^{1}$\quad Keita Ishii$^{2}$\quad Hiroki Sakamoto$^{2}$\quad Manabu Kikuchi$^{2}$\quad Katsuma Kitamura$^{2}$\\
Chiho Semba$^{2}$\quad Kenta Nishiyama$^{2}$\quad Teppei Mori$^{2}$\quad Yukihiro Kiwaki$^{2}$\quad Takeo Kato$^{1}$\thanks{Corresponding author.}\\[3pt]
$^{1}$Keio University, Japan\qquad $^{2}$Bridgestone Corporation, Japan\\[3pt]
{\small\tt tang-jianbo@keio.jp, kato@mech.keio.ac.jp}\\
{\small\tt \{keita.ishii, hiroki.sakamoto2, manabu.kikuchi, katsuma.kitamura,}\\
{\small\tt chiho.semba, kenta.nishiyama1, teppei.mori, yukihiro.kiwaki\}@bridgestone.com}
}

\begin{document}
\maketitle
{\let\thefootnote\relax\footnotetext{\copyright~2027 IEEE. Personal use of this material is permitted. Permission from IEEE must be obtained for all other uses, in any current or future media, including reprinting/republishing this material for advertising or promotional purposes, creating new collective works, for resale or redistribution to servers or lists, or reuse of any copyrighted component of this work in other works.}}

\begin{abstract}
Requested-value control is stricter than continuous steering: given $c$ on a human-rating scale, the generated image should attain that value from only limited ratings. We present \textbf{ScoreDial}, which fits a differentiable regressor $R$ to 100--479 human-rated images; once $R$ is fixed, a requested value $c$ is defined on its output scale, and learned target conditioning is paired with error-adaptive guidance that repeatedly reduces $|R(\mathbf{x})-c|$ during sampling. Across tire slip resistance, interior visual complexity, food desire-to-eat, and soap translucency, using one set of sampler settings, ScoreDial reaches requested values and achieves $R$-MAE at most \ControlMAEMax{} over a range of width 0.8 and Spearman $\rho=\ControlRhoAll$, the maximum attainable under our protocol. Because $R$ also drives sampling, we evaluate agreement with $R$ separately from matched sweeps, an independent auditor fitted to the same ratings on a backbone no deployed regressor uses, and blinded human judgments. MPGD-flow also reaches the requested values under $R$, but ScoreDial has larger auditor displacement on every attribute, the largest among all compared methods, and a stronger MPGD-flow setting gives up target agreement without closing this gap. On every attribute, removing target conditioning leaves low $R$-MAE but more than halves auditor displacement, separating precise score correction from attribute movement. Blinded rating studies with \HumanParticipants{} raters per attribute show mean human ratings track $c$ on tire and soap, with linear $R^2=\HumanTireLinearRtwo$ and $R^2=\HumanSoapLinearRtwo$, and in forced-choice comparisons of matched sweeps, raters chose ScoreDial's as the clearer progression in at least \TwoAFCMinPct\% of judgments against every baseline. Code: \url{https://github.com/Kamijo-Gen/ScoreDial}.
\end{abstract}

\begin{figure*}[tbp]
\centering
\includegraphics[width=0.86\textwidth]{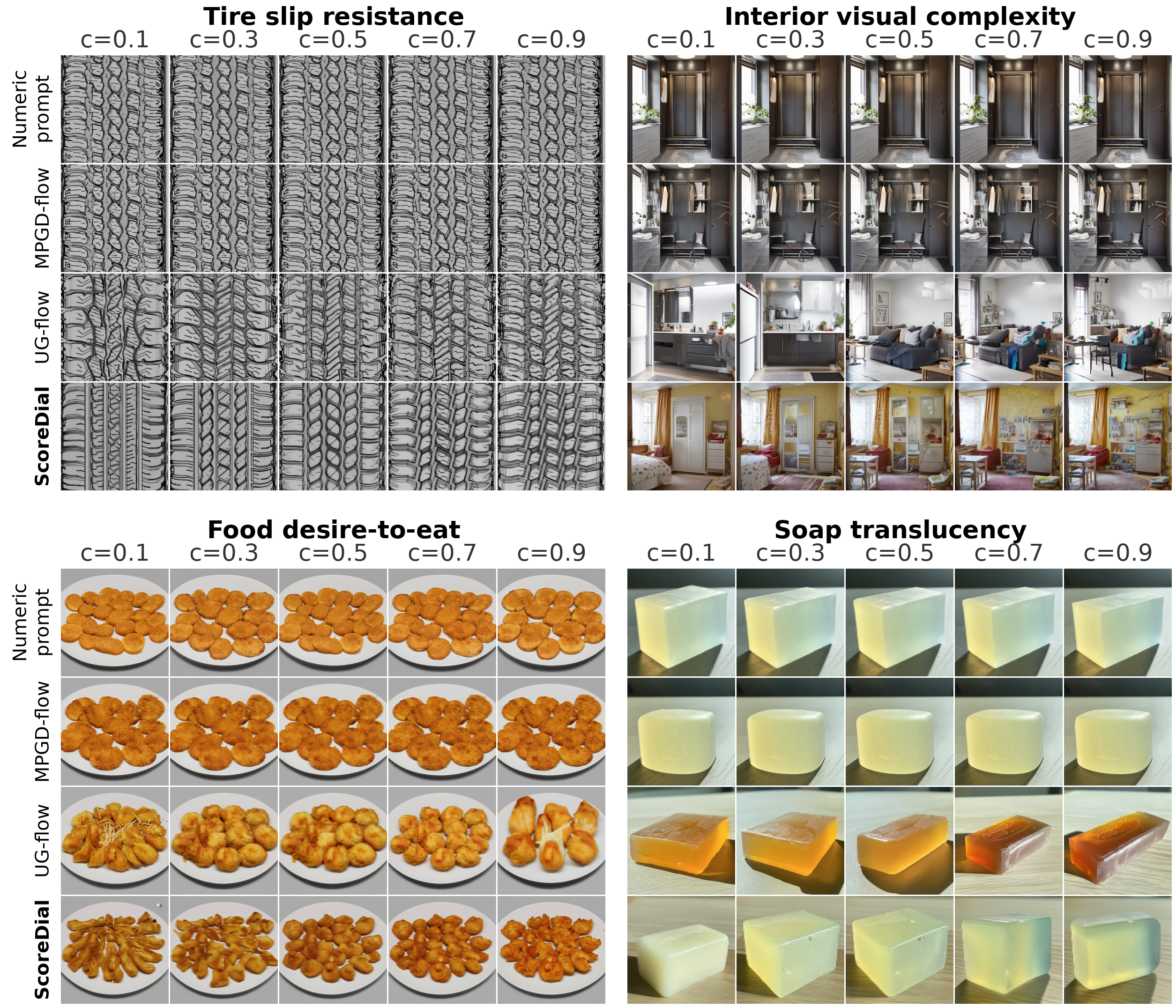}
\caption{Matched requested-value sweeps for the four compared methods (MPGD-flow at its selected strength); $c$ increases from left to right in steps of 0.2. All methods share the frozen SD3.5 model, resolution, base sampler settings, prompts, and---within each attribute---the same seed.}
\label{fig:sweep}
\end{figure*}

\section{Introduction}
\label{sec:intro}
Visual appearance influences how products are perceived and evaluated~\cite{Bloch1995,Crilly2004}. Such judgments concern perceived attributes of an appearance~\cite{Crilly2004}, and collecting consumer evaluations at scale can be expensive~\cite{Burnap2023}. Generative models can automate the creation of design images~\cite{Burnap2023,Yang2023}, but useful control may require more than asking for an attribute to become ``more'' or ``less'' pronounced: the desired outcome can be a particular numeric level~\cite{Chen2026,Cho2025,Ibarrola2024}.

We study \emph{requested-value control} of task-specific human-rated attributes from only a few hundred rated images. Continuous conditional generators are developed in label-rich settings and condition on scalar regression labels~\cite{Ding2021,Ding2024,Ding2026,Cho2025,Zhu2025}, but sampling is open-loop to realized label error: the current sample's discrepancy is not measured and fed back. Slider methods expose a continuous strength or direction~\cite{Ezra2025,Gandikota2024,Baumann2025,Chiu2026,Zarei2026,Ye2026}, not a requested value calibrated to a task-specific human-rating scale. Other controls derive attributes from image statistics, classifiers, or vision--language models~\cite{Chen2026,Zhu2025}; Hu et al. describe CLIPScore as ``unreliable and often inaccurate'' for text--image faithfulness~\cite{Hu2023}. Our setting combines limited human ratings with a numeric request on that scale.

ScoreDial turns those ratings into a closed-loop generator. A frozen pretrained visual backbone and a linear ridge head fitted to the human ratings form the differentiable regressor $R$, while a pretrained diffusion model is adapted to each image domain. \textbf{Target conditioning}, trained jointly with the domain LoRA on $R$-pseudo-labeled images, injects $c$ into every transformer block and positions generation open-loop. \textbf{Error-adaptive regressor guidance} then closes the loop by repeatedly measuring $R(\mathbf{x})-c$ on the clean-image estimate and applying a gradient-normalized, noise-weighted correction that shrinks with the remaining error. We write $R$-MAE for the mean of $|R(\mathbf{x})-c|$ over a stated grid of targets and seeds.

Because $R$ also drives sampling, low $R$-MAE establishes controller agreement but not by itself perceptual change. Training-free loss-based guidance can be more susceptible to adversarial gradients than classifier guidance~\cite{Shen2024}, and learned objectives can be over-optimized~\cite{Gao2023,Clark2024,Zhang2024}. We therefore separate \emph{target agreement under $R$} from \emph{attribute movement}, assessed by matched sweeps, an independent auditor using the same ratings but a backbone absent from all deployed regressors, and blinded human judgments.

\noindent\textbf{Contributions.}
(1)~We formulate and evaluate requested-value generation for task-specific human-rated visual attributes with only 100--479 rated images per attribute.
(2)~We introduce \textbf{ScoreDial}, which couples learned target conditioning with error-adaptive regressor feedback so that generation is first positioned by $c$ and then corrected according to the realized score error.
(3)~Across four disparate attributes, we show empirically that convergence under the guidance regressor $R$ and movement of the intended attribute are not interchangeable: MPGD-flow, and ScoreDial's own feedback without target conditioning, reach the requested values under $R$ but move the attribute less than ScoreDial, which has the largest auditor displacement among the compared methods on all four attributes. Blinded ratings on tire and soap confirm that mean perceived levels track $c$, and in forced-choice judgments raters chose ScoreDial's progression as the clearer one in at least \TwoAFCMinPct\% of comparisons with every baseline.

\section{Related Work}
\label{sec:related}
\noindent\textbf{Human-rated objectives in design and affective generation.}
Human ratings have also shaped design-generation pipelines. Burnap et al. predict consumer aesthetic scores and automatically generate product designs~\cite{Burnap2023}; Hayakawa et al. train StyleGAN on tread images in the top 40\% under a slip-resistance regressor~\cite{Hayakawa2024}; Yang et al. score and filter diffusion-generated products with a Kansei predictor~\cite{Yang2023}. Ibarrola et al. instead target requested valence--arousal--dominance values using an affect predictor trained from 13,913 rated words and 1,194 rated images~\cite{Ibarrola2024}. Their optimization-based generators include affect-prediction error in the objective, while their Stable Diffusion procedure optimizes text-conditioning embeddings for the requested affect. ScoreDial targets task-specific scales learned from a few hundred rated images and repeatedly corrects the realized image score during sampling.

\noindent\textbf{Continuous scalar-conditioned generation.}
CcGAN~\cite{Ding2021,Ding2023}, CCDM~\cite{Ding2024}, and iCCDM~\cite{Ding2026} learn image generators conditioned on continuous scalar regression labels. Their published evaluation protocols use regressors trained on real images to measure label consistency after generation, but their sampling procedures do not feed the predicted label error back to update each sample. Att-Adapter adapts a pretrained diffusion model for continuous multi-attribute control using 16,336 face images, whose attributes come from landmarks and pretrained face estimators, and 2,030 car images~\cite{Cho2025}. CompSlider turns sixteen attribute values, obtained from attribute classifiers over roughly three million training images, into the conditioning a frozen text-to-image model consumes~\cite{Zhu2025}. Our conditioning stage follows the same principle, but its pseudo-labels come from an $R$ fitted to only 100--479 human-rated images rather than the much larger labeled sets these methods use, and it is paired at inference with feedback based on $R(\mathbf{x})-c$.

\noindent\textbf{Intensity controls and sliders.}
AttriCtrl maps concrete attributes and vision--language-derived quantities to a unified $[0,1]$ scale and injects user values through learned encoders with a frozen diffusion backbone~\cite{Chen2026}. Slider methods instead learn or identify continuous directions in model parameters, text embeddings, or latent representations~\cite{Gandikota2024,Baumann2025,Chiu2026,Ezra2025,Zarei2026,Ye2026}; their coefficients are not fitted to ratings of a task-specific human attribute. Across these works, reported attribute change is scored by vision--language models: $\Delta$CLIP~\cite{Gandikota2024,Ezra2025,Baumann2025,Chiu2026}, CLIP/SigLIP/BLIP similarity~\cite{Zarei2026}, or a multimodal language model~\cite{Ye2026}, rather than ratings of that attribute.

\noindent\textbf{Inference-time guidance and reliability.}
Universal Guidance~\cite{Bansal2024} and MPGD~\cite{He2024} inject differentiable objectives during diffusion sampling. Related guidance methods include DPS~\cite{Chung2023}, LGD~\cite{Song2023}, and TFG~\cite{Ye2024}. Gebhardt et al. use a learned regressor to guide image editing~\cite{Gebhardt2025}, whereas ScoreDial generates from noise. Our sampler adapts MPGD-LDM's clean-estimate update and Universal Guidance's stepwise refinement, but makes the correction magnitude depend explicitly on the current absolute score error and pairs this feedback stage with learned target conditioning.

\begin{figure*}[tbp]
\centering
\includegraphics[width=0.95\textwidth]{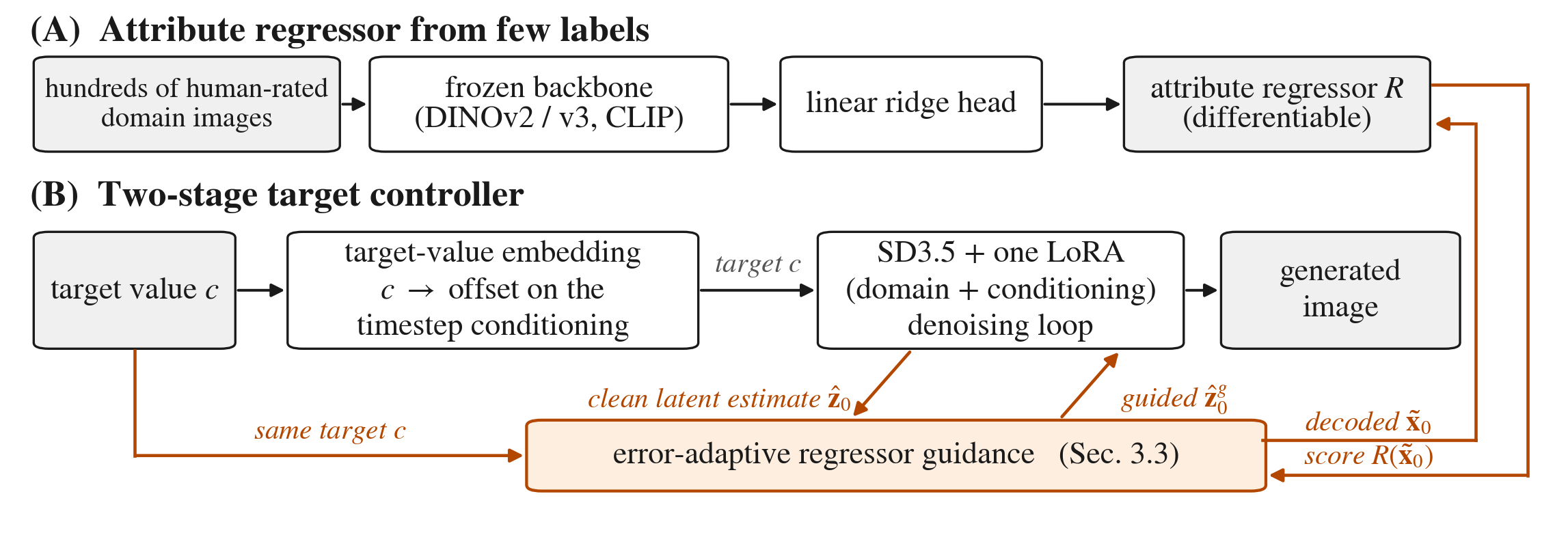}
\caption{ScoreDial separates target positioning from residual correction. \textbf{(A)}~A frozen pretrained visual backbone and a fitted linear ridge head form the differentiable attribute regressor $R$. \textbf{(B)}~A learned embedding of the requested value $c$ is injected into every transformer block and trained jointly with a domain LoRA, providing open-loop target conditioning. During active guidance steps, $R$ evaluates the decoded clean estimate; the squared-error gradient is normalized, and the clean latent is corrected with a step size determined by the current error and noise level.}
\label{fig:pipeline}
\end{figure*}

\section{Method}
\label{sec:method}
ScoreDial first fits a differentiable regressor $R$ to human ratings (Sec.~\ref{sec:reg}), then combines target conditioning that places generation near $c$ (Sec.~\ref{sec:cond}) with error-adaptive guidance that closes the residual gap during sampling (Sec.~\ref{sec:guid}; Fig.~\ref{fig:pipeline}). Both operate on $R$'s scale.

\subsection{Attribute regressor from few labels}
\label{sec:reg}
A frozen pretrained backbone maps each image to $\mathbf{f}\in\mathbb{R}^{d}$, and the human ratings fit a ridge head. We consider seven backbone--representation candidates: for base DINOv2 and DINOv3~\cite{Oquab2023,Simeoni2025} (excluding register tokens), the class token, patch-token mean, and their concatenation; for CLIP ViT-B/16~\cite{Radford2021}, the projected pooled representation. Selection and guidance use identical preprocessing: RGB $[0,1]$, bicubic short-edge resize to 256, $224\times224$ center crop, and pretrained normalization (Sec.~\ref{sec:regeval}).

A ridge-regression head is fitted to standardized features:
\begin{equation}
R(\mathbf{x})=\boldsymbol{\beta}^{\top}\tilde{\mathbf{f}}+b,\qquad \tilde{\mathbf{f}}=(\mathbf{f}-\boldsymbol{\mu}_f)/\boldsymbol{\sigma}_f.
\end{equation}
Here $\boldsymbol{\mu}_f$ and $\boldsymbol{\sigma}_f$ are the per-dimension mean and scale estimated from the training features, and $\boldsymbol{\beta}$ and $b$ are the ridge head parameters. Freezing the feature extractor keeps the learned component small, while the resulting image-to-score function remains end-to-end differentiable, allowing gradients to propagate from $R$ through the decoded clean estimate to the predicted clean latent.

\subsection{Domain adaptation and target conditioning}
\label{sec:cond}
The generator is the official \textbf{Stable Diffusion 3.5-medium} release, whose rectified-flow transformer follows SD3~\cite{Esser2024}.\footnote{\raggedright\url{https://huggingface.co/stabilityai/stable-diffusion-3.5-medium}; we use revision \texttt{b940f670}.} A single LoRA~\cite{Hu2022} adapts the transformer to each image domain and is trained jointly with the target-value embedding below; Sec.~\ref{sec:setup} specifies the fixed training pools.

To place generation according to the requested value, we inject $c$ through the transformer's existing scalar-conditioning pathway. We first standardize it,
\begin{equation}
\bar c=\frac{c-\mu_R}{\sigma_R},
\end{equation}
where $\mu_R$ and $\sigma_R$ are the mean and standard deviation of $R$'s predictions on the domain images (Sec.~\ref{sec:setup}). A sinusoidal projection followed by a two-layer MLP embeds $\bar c$; the result is added to the combined timestep and pooled-text embedding that conditions every transformer block. The same value embedding is supplied to both conditional and unconditional classifier-free-guidance branches, preventing CFG from multiplying the requested value.

The deployed $R$ pseudo-labels every domain-training image, so conditioning and feedback use the same scale. These labels supervise joint flow-matching training of the LoRA and value embedding; no additional human ratings are used. The embedding's last layer is zero-initialized and is present at every denoising step. This path is open-loop: it sees $c$ but not the current sample's score, leaving residual error for feedback.

\subsection{Error-adaptive regressor guidance}
\label{sec:guid}
Target conditioning provides a location on the scale; guidance now corrects the realized sample. We use 30 denoising steps, indexed by $t\in\{0,\ldots,29\}$. At each guidance pass, let $\mathbf{z}$ be the current noisy latent and $\sigma\in(0,1]$ the current noise level. We compute the classifier-free-guidance (CFG) velocity~\cite{Ho2022}, its clean prediction, the decoded RGB estimate, and the signed score error:
\begin{subequations}\label{eq:pass}
\begin{align}
\mathbf{v}&=\mathbf{v}_\theta^{\mathrm{u}}+\omega\bigl(\mathbf{v}_\theta^{\mathrm{c}}-\mathbf{v}_\theta^{\mathrm{u}}\bigr),\label{eq:cfg}\\
\hat{\mathbf{z}}_0&=\mathbf{z}-\sigma \mathbf{v},\label{eq:cleanpred}\\
\tilde{\mathbf{x}}_0&=\frac{\operatorname{clip}\!\left(\mathcal{D}(\hat{\mathbf{z}}_0),-1,1\right)+1}{2},\label{eq:decode}\\
\mathrm{err}&=R(\tilde{\mathbf{x}}_0)-c.\label{eq:err}
\end{align}
\end{subequations}
$\omega$ is the CFG scale; both branches share the target-value embedding (Sec.~\ref{sec:cond}). $\mathcal{D}$ is the frozen Stable Diffusion 3.5 VAE decoder. Clipping and remapping to $[0,1]$ remain inside the computational graph and are followed by the resize, crop, and model-specific normalization of Sec.~\ref{sec:reg}; on the rated images, this frozen autoencoder path alone changes $R$ by at most \RoundTripPctMax\% of the control range (supplementary material). We then differentiate the squared target error with respect to the predicted clean latent:
\begin{subequations}\label{eq:gradient}
\begin{align}
\mathbf{g}&=\nabla_{\hat{\mathbf{z}}_0}\mathrm{err}^{2}
=2\,\mathrm{err}\,\nabla_{\hat{\mathbf{z}}_0}R(\tilde{\mathbf{x}}_0),\label{eq:gradient-raw}\\
\hat{\mathbf{g}}&=\frac{\mathbf{g}}{\lVert \mathbf{g}\rVert+10^{-8}},\label{eq:gradient-unit}
\end{align}
\end{subequations}
where the norm is computed independently for each latent. We use the gradient for direction but not for step magnitude: after normalization, the guided clean prediction is
\begin{equation}
\hat{\mathbf{z}}_0^{\,g}=\hat{\mathbf{z}}_0-s_t\lVert\hat{\mathbf{z}}_0\rVert\hat{\mathbf{g}}.
\end{equation}
The correction magnitude is set explicitly by the current error and noise level:
\begin{subequations}\label{eq:adaptive-step}
\begin{align}
s_t&=\lambda_t|\mathrm{err}|\,w(\sigma),\label{eq:step-size}\\
w(\sigma)&=w_{\min}+(1-w_{\min})\sigma^q.\label{eq:noise-weight}
\end{align}
\end{subequations}
This form makes the feedback self-attenuating: $s_t$ is proportional to $|R(\tilde{\mathbf{x}}_0)-c|$ and approaches zero as the requested value is reached. Gradient normalization prevents the raw norm of $\nabla R$ from determining the update size, while $\lVert\hat{\mathbf{z}}_0\rVert$ expresses the displacement relative to the clean-latent magnitude. Following guidance scaling~\cite{Dhariwal2021} and timestep-dependent scaling in Universal Guidance~\cite{Bansal2024}, $w(\sigma)$ weakens the correction as the sample becomes cleaner while retaining a floor $w_{\min}$.

With normalized progress $p=t/30$, the guidance coefficient is
\begin{equation}
\lambda_t=
\begin{cases}
0, & p<0.2,\\
\lambda_{\mathrm{main}}, & 0.2\leq p\leq0.8,\\
\lambda_{\mathrm{tail}}, & p>0.8.
\end{cases}
\end{equation}
The first 20\% of the trajectory therefore uses target-conditioned CFG without regressor guidance, and the error factor and noise weighting remain active in both guidance intervals. We set $\lambda_{\mathrm{main}}=0.4$, $\lambda_{\mathrm{tail}}=0.05$, $w_{\min}=0.2$, and $q=1$ once in preliminary tire experiments and reuse them unchanged for all four attributes; no sampler setting is tuned per attribute.

\noindent\textbf{Sampling transition and refinement.}
We apply the correction to the predicted clean latent rather than back-propagating through the denoiser, following the clean-estimate update in MPGD-LDM~\cite{He2024} (Algorithm~2). The current velocity also defines the corresponding noise prediction
\begin{equation}
\hat{\boldsymbol{\epsilon}}=\mathbf{z}+(1-\sigma)\mathbf{v}.
\end{equation}
After the final guidance pass at a denoising step, the next rectified-flow state is
\begin{equation}
\mathbf{z}_{\mathrm{next}}=(1-\sigma_{\mathrm{next}})\hat{\mathbf{z}}_0^{\,g}
+\sigma_{\mathrm{next}}\hat{\boldsymbol{\epsilon}}.
\end{equation}
Here $\sigma_{\mathrm{next}}$ is the next schedule value. For every step with $\lambda_t>0$, we additionally adapt Universal Guidance's Universal Stepwise Refinement~\cite{Bansal2024}. We use $k=2$ guidance passes in total. Between the two passes, the state is re-noised to a level $\sigma_{\mathrm{back}}$ as far above $\sigma$ as $\sigma_{\mathrm{next}}$ is below it:
\begin{equation}
\begin{aligned}
\sigma_{\mathrm{back}}&=\sigma+(\sigma-\sigma_{\mathrm{next}}),
\qquad
\kappa=\frac{1-\sigma_{\mathrm{back}}}{1-\sigma_{\mathrm{next}}},\\[4pt]
\mathbf{z}&\leftarrow\kappa\,\mathbf{z}_{\mathrm{next}}
+\sqrt{\sigma_{\mathrm{back}}^{2}-(\kappa\,\sigma_{\mathrm{next}})^{2}}\;\boldsymbol{\epsilon},
\enspace \boldsymbol{\epsilon}\sim\mathcal{N}(0,\mathbf{I}).
\end{aligned}
\end{equation}
Because $\kappa\mathbf{z}_{\mathrm{next}}=(1-\sigma_{\mathrm{back}})\hat{\mathbf{z}}_0^{\,g}+\kappa\sigma_{\mathrm{next}}\hat{\boldsymbol{\epsilon}}$, the estimated noise is carried into the refinement state and only the remaining noise is sampled afresh. The second pass recomputes the velocity, clean estimate, score error, and gradient at $\sigma_{\mathrm{back}}$, using the nearest schedule timestep for denoiser conditioning; the final transition still advances to this step's $\sigma_{\mathrm{next}}$. Steps with $\lambda_t=0$ use a single target-conditioned CFG Euler transition with neither regressor guidance nor refinement.

\section{Experiments}
\label{sec:exp}
We test whether ScoreDial reaches requested values and moves the corresponding attribute (Sec.~\ref{sec:control}), which components supply precision and movement (Sec.~\ref{sec:ablation}), and whether people perceive the requested progression, also relative to the baselines (Sec.~\ref{sec:human}).

\subsection{Setup}
\label{sec:setup}
\noindent\textbf{Source supervision.}
We evaluate four attributes with different semantics and rating protocols. \textbf{Tire slip resistance} uses five-point Likert ratings from Hayakawa et al.~\cite{Hayakawa2024}. Their 120 tread entries contain one duplicated image with conflicting means; dropping both entries leaves \textbf{118 rated images}. \textbf{Interior visual complexity} uses the complete 100-image Interior Design category of SAVOIAS~\cite{Saraee2018}, whose 0--100 scores were derived by the dataset authors from crowdsourced pairwise comparisons using Bradley--Terry and matrix completion. \textbf{Food desire-to-eat} uses the \textbf{479 CROCUFID images} with 0--100 visual-analogue ratings~\cite{Toet2019}. \textbf{Soap translucency} uses the \textbf{150 real photographs} rated on seven-point scales by Liao et al.~\cite{Liao2023}, excluding their 150 model-generated soap images; following their protocol, we min--max normalize each observer's ratings and then average across observers.

\noindent\textbf{Generator-training images.}
Domain adaptation and target conditioning share one image pool per attribute: \textbf{984} tires, \textbf{100} interiors, \textbf{479} foods, and \textbf{800} soaps. The tire pool begins with all 534 unique tread images in the patent-sourced archive of Hayakawa et al.~\cite{Hayakawa2024}; the 535-file archive contains one exact byte-level duplicate, and 118 unique images carry ratings. For generator training only, the original patent line drawings of these treads are upsampled to $1024^2$ with Topaz Gigapixel, median-filtered with a $7\times7$ window, and standardized per image to mean 128 and standard deviation 55, clipped to $[0,255]$. $R$ is always fitted to the rated archive images, and no upsampled image enters a reported measurement. To reduce rendering artifacts observed at extreme-score tire targets, we oversample the score tails deterministically: using deployed-$R$ scores of the archive images, horizontal, vertical, and both-axis flips are added for every image with score $\leq0.252$ (65 images) or $\geq0.746$ (85 images), producing 984 training images. Interior and food use exactly their rated sets for generator training. The 800 soaps are sampled uniformly at random from the 8,085-photograph Translucent Image Dataset~\cite{Liao2023} before scoring. An SSCD copy-detection check~\cite{Pizzi2022} finds no reported image unusually close to its generator-training pool (supplementary material).

\noindent\textbf{Score ranges and requested targets.}
All requested values are represented on $[0,1]$, so the error factor $|R(\mathbf{x})-c|$ has a common scale across attributes: the observed tire and food label ranges are mapped affinely, interior's defined 0--100 scale by its endpoints, and the averaged soap scores already lie in $[0,1]$. Every control experiment and ablation uses the same five targets, $c\in\{0.1,0.3,0.5,0.7,0.9\}$, with ten seeds per target.

\noindent\textbf{Training and sampling.}
We use Stable Diffusion 3.5-medium through Hugging Face diffusers in PyTorch on one NVIDIA RTX 5090. The fixed domain prompts are ``a tire tread pattern, top-down view, grayscale,'' ``a photograph of a room interior,'' ``a plate of food on white background,'' and ``a bar of soap.'' For tire/interior/food/soap respectively, a rank-32 LoRA on the transformer's attention projections and the target-value embedding are trained jointly for 3,000/3,000/2,000/6,000 flow-matching steps, with batch 4, AdamW at constant learning rate $10^{-4}$, and horizontal-flip probability 0.5. Training and all reported generation use $1024^2$ images. ScoreDial uses 30 Euler steps, CFG 5, and the guidance schedule of Sec.~\ref{sec:guid}. ``$\pm$'' denotes mean $\pm$ population SD over the stated runs or seeds.

\subsection{Attribute regressors: selection and evaluation}
\label{sec:regeval}
Because the same labeled set is used to choose a representation and to fit the final regressor, we separate evaluation of the \emph{selection procedure} from selection of the \emph{deployment model}~\cite{Cawley2010}. Repeated nested cross-validation of the full selection procedure (supplementary material) gives tire $R^2=\RegTireMean\pm\RegTireStd$, interior $R^2=\RegInteriorMean\pm\RegInteriorStd$, food $R^2=\RegFoodMean\pm\RegFoodStd$, and soap $R^2=\RegSoapMean\pm\RegSoapStd$. For deployment, repeated 5-fold CV on all labels selects DINOv2 patch-token mean with $\alpha=50$ for tire, CLIP projected pooled output with $\alpha=100$ for interior and food, and DINOv3 class-token plus patch-token-mean concatenation with $\alpha=100$ for soap; each head is refit on all labels and fixed before generation begins.

\subsection{Requested-value control: agreement and movement}
\label{sec:control}
We evaluate two axes because $R$ participates in generation: \emph{target agreement}, whether $R$ scores the image at the requested value, and \emph{attribute movement}, whether changing the request changes the attribute under evidence the sampler never optimizes.

\noindent\textbf{ScoreDial target agreement.}
A requested value is counted as reached when $|R(\mathbf{x})-c|\leq\ReachTolerance{}$. ScoreDial reaches every tested cell. Across the \ControlCells{} target--seed cells, the median error is at most \ControlMedianMax{}, the 90th percentile at most \ControlPNinetyMax{}, and the largest single-cell error is \ControlWorstCell{} on \ControlWorstAttribute{}. The requested variable has five tied values while generated scores are distinct, so Spearman correlation cannot exceed \ControlRhoCeiling{} on this grid; ScoreDial reaches that ceiling on all four attributes, with no target cell ranked out of order.

\noindent\textbf{Matched comparison protocol.}
Numeric prompt, MPGD-flow, UG-flow, and ScoreDial use the same frozen SD3.5 model, $1024^2$ resolution, 30-step Euler base schedule, CFG 5, fixed domain prompts, and test seeds. Baseline LoRAs are trained exactly as ScoreDial's except that they do not receive the target-value embedding. MPGD-flow and UG-flow use $R$ only during sampling; Numeric prompt uses no regressor guidance. With batch size 1 on the RTX 5090 of Sec.~\ref{sec:setup}, per-image sampling time averaged over the ten test seeds and the four attributes is 5.5~s for Numeric prompt, 16.6~s for MPGD-flow, 48.2~s for UG-flow, and 29.5~s for ScoreDial, at peak memory 7.7/11.9/26.0/11.9~GB; per-attribute values differ from these means by at most 2\%.

Numeric prompt appends ``with perceived [attribute] rated $c$ on a zero-to-one scale,'' using the mapped scales of Sec.~\ref{sec:setup}. MPGD-flow adapts MPGD-LDM Algorithm~2~\cite{He2024} to rectified flow: one fixed-coefficient loss-gradient step is applied to the predicted clean latent, with neither refinement nor denoiser back-propagation. UG-flow adapts forward Universal Guidance~\cite{Bansal2024} to rectified flow and differentiates through the denoiser. It uses no backward optimization steps, matching Universal Guidance's Stable Diffusion setting, and uses Universal Stepwise Refinement with the same $k=2$ as ScoreDial.

The continuous scalar-conditioned generators discussed in Sec.~\ref{sec:related} train their own scalar-conditioned models rather than operating as control mechanisms on the matched frozen SD3.5 backbone~\cite{Ding2021,Ding2024,Ding2026}; both these and the closest methods built on pretrained diffusion models~\cite{Cho2025,Zhu2025} report published settings substantially more label-rich than the 100--479 human-rated images available here. We therefore do not include them in this comparison. The conditioning-only ablation (Sec.~\ref{sec:ablation}) additionally evaluates the same open-loop scalar-conditioning principle under the matched protocol on every attribute.

\noindent\textbf{Guidance-strength selection.}
Strengths are chosen using ten seeds disjoint from all test seeds. For each guidance baseline we search $\{1,3,10,30,100,300\}$ at endpoints 0.1 and 0.9 and choose the lowest endpoint $R$-MAE among settings that still produce images of the domain; endpoint selection yields one fixed coefficient for the full requested range. MPGD-flow selects 100/30/30/10 for tire/interior/food/soap, each strictly inside the ladder, with error increasing on both neighboring sides. UG-flow's error decreases at every rung, but strengths from 100 upward produce hard-edged color blocks rather than domain images for every attribute. We therefore report UG-flow at 30, the strongest domain-valid rung and hence the lowest-error eligible setting. This rule optimizes target agreement alone. UG-flow's selected rung is already its strongest domain-valid one, whereas MPGD-flow's lies inside the ladder, so a stronger setting could trade agreement for movement. We thus also report MPGD-flow one rung stronger, at \MPGDStrongTire/\MPGDStrongInterior/\MPGDStrongFood/\MPGDStrongSoap{} (MPGD-flow$^{+}$), a setting less restrictive in the precision/movement trade-off.

\begin{table*}[!t]
\centering\footnotesize
\setlength{\tabcolsep}{3pt}
\caption{Matched comparison. Ten test seeds are shared by all methods and all requested values ($N{=}\ExtTireOursN$ images per method and attribute); MPGD-flow$^{+}$ is MPGD-flow one strength rung above its selected one (Sec.~\ref{sec:control}). \textbf{Top:} agreement with the deployed regressor $R$; $\pm$ is the SD of the per-seed mean, and $\rho$ is Spearman correlation between requested value and $R$ of the generated image. \textbf{Bottom:} independent-auditor endpoint displacement $A(0.9)-A(0.1)$ within one seed (mean $\pm$ SD over the \AudSeeds{} seeds); auditors are attribute-specific, so comparisons are within a row.}
\label{tab:external}
\begin{tabular}{lcccccccccc}
\toprule
& \multicolumn{2}{c}{Numeric prompt} & \multicolumn{2}{c}{MPGD-flow} & \multicolumn{2}{c}{MPGD-flow$^{+}$}
& \multicolumn{2}{c}{UG-flow} & \multicolumn{2}{c}{ScoreDial} \\
\cmidrule(lr){2-3}\cmidrule(lr){4-5}\cmidrule(lr){6-7}\cmidrule(lr){8-9}\cmidrule(lr){10-11}
Attribute & $R$-MAE$\downarrow$ & $\rho\uparrow$ & $R$-MAE$\downarrow$ & $\rho\uparrow$ & $R$-MAE$\downarrow$ & $\rho\uparrow$
& $R$-MAE$\downarrow$ & $\rho\uparrow$ & $R$-MAE$\downarrow$ & $\rho\uparrow$ \\
\midrule
Tire & $\ExtTirePromptMAE\!\pm\!\ExtTirePromptMAESD$ & $\ExtTirePromptRho$ & $\ExtTireMPGDMAE\!\pm\!\ExtTireMPGDMAESD$ & $\ExtTireMPGDRho$ & $\ExtTireMPGDsMAE\!\pm\!\ExtTireMPGDsMAESD$ & $\ExtTireMPGDsRho$ & $\ExtTireUGMAE\!\pm\!\ExtTireUGMAESD$ & $\ExtTireUGRho$ & $\ExtTireOursMAE\!\pm\!\ExtTireOursMAESD$ & $\ExtTireOursRho$ \\
Interior & $\ExtInteriorPromptMAE\!\pm\!\ExtInteriorPromptMAESD$ & $\ExtInteriorPromptRho$ & $\ExtInteriorMPGDMAE\!\pm\!\ExtInteriorMPGDMAESD$ & $\ExtInteriorMPGDRho$ & $\ExtInteriorMPGDsMAE\!\pm\!\ExtInteriorMPGDsMAESD$ & $\ExtInteriorMPGDsRho$ & $\ExtInteriorUGMAE\!\pm\!\ExtInteriorUGMAESD$ & $\ExtInteriorUGRho$ & $\ExtInteriorOursMAE\!\pm\!\ExtInteriorOursMAESD$ & $\ExtInteriorOursRho$ \\
Food & $\ExtFoodPromptMAE\!\pm\!\ExtFoodPromptMAESD$ & $\ExtFoodPromptRho$ & $\ExtFoodMPGDMAE\!\pm\!\ExtFoodMPGDMAESD$ & $\ExtFoodMPGDRho$ & $\ExtFoodMPGDsMAE\!\pm\!\ExtFoodMPGDsMAESD$ & $\ExtFoodMPGDsRho$ & $\ExtFoodUGMAE\!\pm\!\ExtFoodUGMAESD$ & $\ExtFoodUGRho$ & $\ExtFoodOursMAE\!\pm\!\ExtFoodOursMAESD$ & $\ExtFoodOursRho$ \\
Soap & $\ExtSoapPromptMAE\!\pm\!\ExtSoapPromptMAESD$ & $\ExtSoapPromptRho$ & $\ExtSoapMPGDMAE\!\pm\!\ExtSoapMPGDMAESD$ & $\ExtSoapMPGDRho$ & $\ExtSoapMPGDsMAE\!\pm\!\ExtSoapMPGDsMAESD$ & $\ExtSoapMPGDsRho$ & $\ExtSoapUGMAE\!\pm\!\ExtSoapUGMAESD$ & $\ExtSoapUGRho$ & $\ExtSoapOursMAE\!\pm\!\ExtSoapOursMAESD$ & $\ExtSoapOursRho$ \\
\midrule
& \multicolumn{10}{c}{Independent-auditor endpoint displacement $A(0.9)-A(0.1)\uparrow$} \\
\midrule
Tire & \multicolumn{2}{c}{$\AudTirePrompt\!\pm\!\AudTirePromptSD$} & \multicolumn{2}{c}{$\AudTireMPGD\!\pm\!\AudTireMPGDSD$} & \multicolumn{2}{c}{$\AudTireMPGDs\!\pm\!\AudTireMPGDsSD$} & \multicolumn{2}{c}{$\AudTireUG\!\pm\!\AudTireUGSD$} & \multicolumn{2}{c}{$\AudTireOurs\!\pm\!\AudTireOursSD$} \\
Interior & \multicolumn{2}{c}{$\AudInteriorPrompt\!\pm\!\AudInteriorPromptSD$} & \multicolumn{2}{c}{$\AudInteriorMPGD\!\pm\!\AudInteriorMPGDSD$} & \multicolumn{2}{c}{$\AudInteriorMPGDs\!\pm\!\AudInteriorMPGDsSD$} & \multicolumn{2}{c}{$\AudInteriorUG\!\pm\!\AudInteriorUGSD$} & \multicolumn{2}{c}{$\AudInteriorOurs\!\pm\!\AudInteriorOursSD$} \\
Food & \multicolumn{2}{c}{$\AudFoodPrompt\!\pm\!\AudFoodPromptSD$} & \multicolumn{2}{c}{$\AudFoodMPGD\!\pm\!\AudFoodMPGDSD$} & \multicolumn{2}{c}{$\AudFoodMPGDs\!\pm\!\AudFoodMPGDsSD$} & \multicolumn{2}{c}{$\AudFoodUG\!\pm\!\AudFoodUGSD$} & \multicolumn{2}{c}{$\AudFoodOurs\!\pm\!\AudFoodOursSD$} \\
Soap & \multicolumn{2}{c}{$\AudSoapPrompt\!\pm\!\AudSoapPromptSD$} & \multicolumn{2}{c}{$\AudSoapMPGD\!\pm\!\AudSoapMPGDSD$} & \multicolumn{2}{c}{$\AudSoapMPGDs\!\pm\!\AudSoapMPGDsSD$} & \multicolumn{2}{c}{$\AudSoapUG\!\pm\!\AudSoapUGSD$} & \multicolumn{2}{c}{$\AudSoapOurs\!\pm\!\AudSoapOursSD$} \\
\bottomrule
\end{tabular}
\end{table*}

\noindent\textbf{What target agreement alone says.}
Numeric prompt has the largest $R$-MAE on every attribute and near-zero rank correlation. UG-flow and MPGD-flow$^{+}$ improve on it but remain above the \ReachTolerance{} tolerance on all four, while ScoreDial has lower $R$-MAE than both in every domain. More revealingly, applying the same tolerance to the grid mean, MPGD-flow and ScoreDial are within it on all four, while the other three settings are not. Under $R$ alone, MPGD-flow and ScoreDial therefore both appear to solve requested-value control.

\noindent\textbf{Independent attribute movement.}
For each attribute we fit an independent auditor: a ridge head on frozen \AudBackbone{} features~\cite{Zhai2023}, a backbone none of the deployed regressors uses. It uses the same human ratings as $R$, holding the attribute definition fixed while changing representation and predictor, and is never used for generator training, sampling, or strength selection. Each auditor attains nested-CV $R^2\ge\AudRtwoMin{}$, following the practice of evaluating generated samples with a predictor trained apart from the generator~\cite{Ding2021,Ding2024}.

Because sampling explicitly closes the error under $R$, not under the auditor, we do not treat the auditor as a second calibration target or compare its absolute reading $A$ directly with $c$. Instead, for each seed we measure how far that reading travels across the requested range, $A(0.9)-A(0.1)$, with both endpoint images generated from the same noise.

The matched sweeps change the interpretation: Numeric prompt and MPGD-flow vary little between endpoint requests, whereas ScoreDial progresses across the range (Fig.~\ref{fig:sweep}; all test images in the supplementary material). ScoreDial also has larger auditor displacement than MPGD-flow on every attribute (Table~\ref{tab:external}, bottom). MPGD-flow thus exposes the central failure mode: reaching the requested value under $R$ does not imply comparable movement under an independent predictor of the same human-rated attribute. A stronger setting does not close this gap: MPGD-flow$^{+}$ has larger auditor displacement than MPGD-flow on every attribute but gives up target agreement (Table~\ref{tab:external}), and its displacement remains below ScoreDial's on all four.

UG-flow shows the complementary failure mode: substantial movement without reaching the requested values under $R$. Its interior sweep also changes scene content between neighboring targets---washbasin, kitchen, living room---whereas ScoreDial keeps one room and changes its complexity. Together, the two panels of Table~\ref{tab:external} show that requested-value control needs both target agreement and attribute movement; ScoreDial is the only compared method that reaches the requested values on all four attributes and has the largest auditor displacement on all four.

\subsection{Ablation of ScoreDial}
\label{sec:ablation}
We ablate ScoreDial on all four attributes to isolate the roles of its two control paths and the design of its feedback update. \emph{Full} is ScoreDial as proposed. \emph{G-only} removes the target-value offset while retaining the same jointly trained LoRA, so $c$ enters only through regressor guidance.

\begin{table*}[!t]
\centering\scriptsize
\setlength{\tabcolsep}{2.5pt}
\caption{Ablation on all four attributes: $R$-agreement and independent-auditor endpoint displacement $\Delta$ on the same $5\!\times\!10$ target--seed grid ($N{=}50$ per configuration and attribute). Full is the ScoreDial run of Table~\ref{tab:external}.}
\label{tab:ablation}
\begin{tabular}{lcccccccc}
\toprule
& \multicolumn{2}{c}{Tire} & \multicolumn{2}{c}{Interior} & \multicolumn{2}{c}{Food} & \multicolumn{2}{c}{Soap} \\
\cmidrule(lr){2-3}\cmidrule(lr){4-5}\cmidrule(lr){6-7}\cmidrule(lr){8-9}
Configuration & $R$-MAE$\downarrow$ & $\Delta\uparrow$ & $R$-MAE$\downarrow$ & $\Delta\uparrow$ & $R$-MAE$\downarrow$ & $\Delta\uparrow$ & $R$-MAE$\downarrow$ & $\Delta\uparrow$ \\
\midrule
Full & $\AblationGridFullMAE\!\pm\!\AblationGridFullMAESD$ & $\AblationAudFull\!\pm\!\AblationAudFullSD$ & $\ExtInteriorOursMAE\!\pm\!\ExtInteriorOursMAESD$ & $\AudInteriorOurs\!\pm\!\AudInteriorOursSD$ & $\ExtFoodOursMAE\!\pm\!\ExtFoodOursMAESD$ & $\AudFoodOurs\!\pm\!\AudFoodOursSD$ & $\ExtSoapOursMAE\!\pm\!\ExtSoapOursMAESD$ & $\AudSoapOurs\!\pm\!\AudSoapOursSD$ \\
G-only & $\AblationGridGOnlyMAE\!\pm\!\AblationGridGOnlyMAESD$ & $\AblationAudGOnly\!\pm\!\AblationAudGOnlySD$ & $\AblInteriorGOnlyMAE\!\pm\!\AblInteriorGOnlyMAESD$ & $\AblInteriorGOnlyAud\!\pm\!\AblInteriorGOnlyAudSD$ & $\AblFoodGOnlyMAE\!\pm\!\AblFoodGOnlyMAESD$ & $\AblFoodGOnlyAud\!\pm\!\AblFoodGOnlyAudSD$ & $\AblSoapGOnlyMAE\!\pm\!\AblSoapGOnlyMAESD$ & $\AblSoapGOnlyAud\!\pm\!\AblSoapGOnlyAudSD$ \\
Conditioning only & $\AblationCondOnlyMAE\!\pm\!\AblationCondOnlyMAESD$ & $\AblationAudCondOnly\!\pm\!\AblationAudCondOnlySD$ & $\AblInteriorCondOnlyMAE\!\pm\!\AblInteriorCondOnlyMAESD$ & $\AblInteriorCondOnlyAud\!\pm\!\AblInteriorCondOnlyAudSD$ & $\AblFoodCondOnlyMAE\!\pm\!\AblFoodCondOnlyMAESD$ & $\AblFoodCondOnlyAud\!\pm\!\AblFoodCondOnlyAudSD$ & $\AblSoapCondOnlyMAE\!\pm\!\AblSoapCondOnlyMAESD$ & $\AblSoapCondOnlyAud\!\pm\!\AblSoapCondOnlyAudSD$ \\
Direct raw gradient & $\AblationRawMAE\!\pm\!\AblationRawMAESD$ & $\AblationAudRaw\!\pm\!\AblationAudRawSD$ & $\AblInteriorRawMAE\!\pm\!\AblInteriorRawMAESD$ & $\AblInteriorRawAud\!\pm\!\AblInteriorRawAudSD$ & $\AblFoodRawMAE\!\pm\!\AblFoodRawMAESD$ & $\AblFoodRawAud\!\pm\!\AblFoodRawAudSD$ & $\AblSoapRawMAE\!\pm\!\AblSoapRawMAESD$ & $\AblSoapRawAud\!\pm\!\AblSoapRawAudSD$ \\
No $|R(\mathbf{x})-c|$ factor & $\AblationConstantMAE\!\pm\!\AblationConstantMAESD$ & $\AblationAudConstant\!\pm\!\AblationAudConstantSD$ & $\AblInteriorConstantMAE\!\pm\!\AblInteriorConstantMAESD$ & $\AblInteriorConstantAud\!\pm\!\AblInteriorConstantAudSD$ & $\AblFoodConstantMAE\!\pm\!\AblFoodConstantMAESD$ & $\AblFoodConstantAud\!\pm\!\AblFoodConstantAudSD$ & $\AblSoapConstantMAE\!\pm\!\AblSoapConstantMAESD$ & $\AblSoapConstantAud\!\pm\!\AblSoapConstantAudSD$ \\
\bottomrule
\end{tabular}
\end{table*}

\begin{figure}[!t]
\centering
\includegraphics[width=0.9\linewidth]{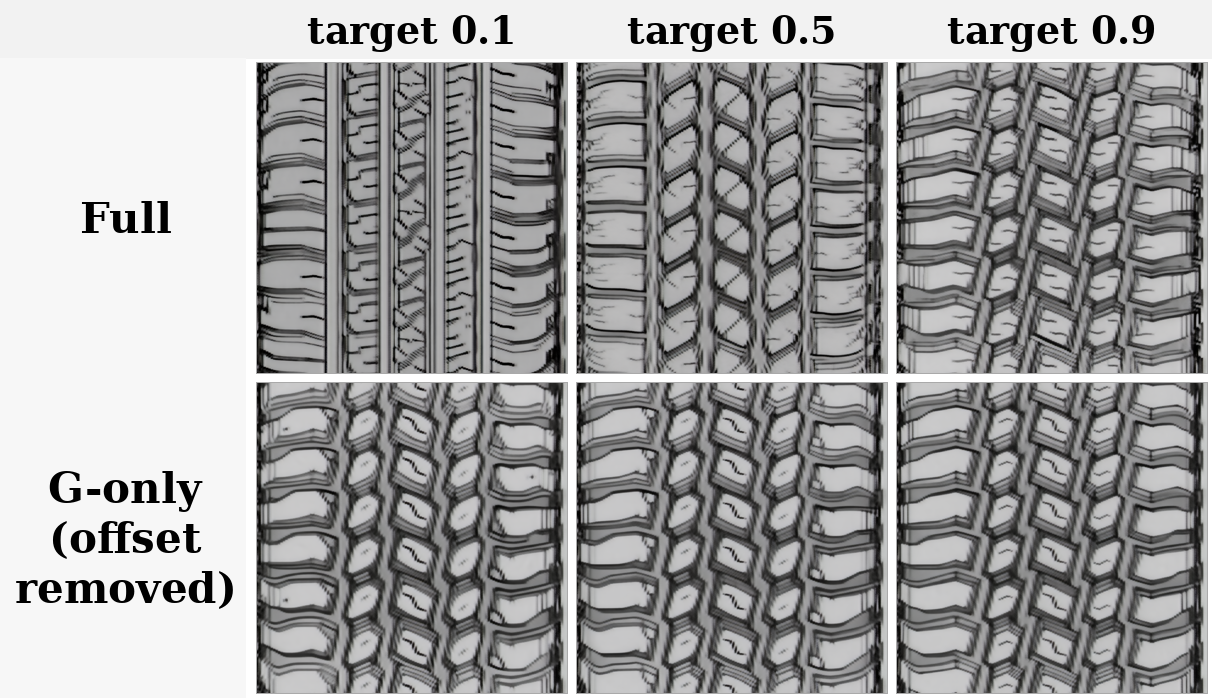}
\caption{Matched Full and G-only tire outputs at requested values 0.1/0.5/0.9; these are the only two configurations that retain low error under $R$ on every attribute (Table~\ref{tab:ablation}).}
\label{fig:tire-ablation}
\end{figure}

\emph{No $|R(\mathbf{x})-c|$ factor} replaces Eq.~\eqref{eq:step-size} by $s_t=\lambda_t w(\sigma)$ while retaining gradient normalization, latent-norm scaling, and refinement. \emph{Direct raw gradient} uses $\hat{\mathbf{z}}_0^{\,g}=\hat{\mathbf{z}}_0-\lambda_t\mathbf{g}$, retaining $\lambda_t$ and refinement but removing gradient normalization, latent-norm scaling, and $w(\sigma)$. \emph{Conditioning only} keeps the target-value embedding and disables regressor guidance.

The update design matters for target precision (Table~\ref{tab:ablation}). On every attribute, removing only the $|R(\mathbf{x})-c|$ factor raises $R$-MAE from Full's \AblAllFullMAELo--\AblAllFullMAEHi{} to \AblAllConstantMAELo--\AblAllConstantMAEHi, and stepping along the raw gradient raises it further, to \AblAllRawMAELo--\AblAllRawMAEHi. Conditioning alone has $R$-MAE \AblAllCondOnlyMAELo--\AblAllCondOnlyMAEHi.

G-only is the only ablation besides Full with low $R$-MAE on every attribute (\AblAllGOnlyMAELo--\AblAllGOnlyMAEHi). Feedback alone can therefore drive $R$ close to $c$ (Fig.~\ref{fig:tire-ablation}; all attributes and configurations in the supplementary material), yet removing target conditioning more than halves auditor displacement on every attribute. Conditioning-only gives auditor displacement of \AblAllCondOnlyAudLo--\AblAllCondOnlyAudHi{} but poor $R$-agreement. The two paths thus separate: target conditioning contributes movement, while feedback supplies precise correction. G-only reproduces the MPGD-flow pattern---low error under $R$ is not sufficient evidence of attribute movement.

\FloatBarrier

\subsection{Human evaluation}
\label{sec:human}
We finally ask whether direct human perception follows the requested value. We run the same blinded protocol on tire and soap. For each attribute, \HumanParticipants{} raters score \HumanImages{} ScoreDial images (five requested values $\times$ eight seeds) without seeing the target. Ratings use the source scale for that attribute: \HumanTirePoints{} points for tire slip resistance~\cite{Hayakawa2024} (1: looks easy to slip; 5: looks difficult to slip) and \HumanSoapPoints{} points for soap translucency~\cite{Liao2023} (1: low; 7: high). Each participant sees \HumanPresentations{} randomized presentations including \HumanRepeats{} blinded repeats.

Participation is voluntary with informed consent; identifiers are pseudonymized, and no demographic or sensitive data are collected. $R$ is never shown to raters and plays no role in rating collection. Before analysis, a rater would be excluded for using at most two distinct values or for mean absolute repeat discrepancy above two points; no rater meets either criterion in either study. Repeated presentations are averaged within rater and then across raters for each image. Mean human ratings track $c$ in both studies: tire gives linear $R^2=\HumanTireLinearRtwo$ and tie-corrected Spearman $\rho=\HumanTireSpearman$, while soap gives $R^2=\HumanSoapLinearRtwo$ and $\rho=\HumanSoapSpearman$.

\begin{table}[!t]
\centering\footnotesize
\setlength{\tabcolsep}{4pt}
\caption{2AFC: fraction of judgments preferring ScoreDial's sweep over the baseline's, with 95\% percentile bootstrap intervals over raters and seeds (0.5 means no preference).}
\label{tab:2afc}
\begin{tabular}{lcc}
\toprule
ScoreDial vs. & Tire & Soap \\
\midrule
Numeric prompt & \TwoAFCTirePrompt{} [\TwoAFCTirePromptLo, \TwoAFCTirePromptHi] & \TwoAFCSoapPrompt{} [\TwoAFCSoapPromptLo, \TwoAFCSoapPromptHi] \\
MPGD-flow & \TwoAFCTireMPGD{} [\TwoAFCTireMPGDLo, \TwoAFCTireMPGDHi] & \TwoAFCSoapMPGD{} [\TwoAFCSoapMPGDLo, \TwoAFCSoapMPGDHi] \\
MPGD-flow$^{+}$ & \TwoAFCTireMPGDs{} [\TwoAFCTireMPGDsLo, \TwoAFCTireMPGDsHi] & \TwoAFCSoapMPGDs{} [\TwoAFCSoapMPGDsLo, \TwoAFCSoapMPGDsHi] \\
UG-flow & \TwoAFCTireUG{} [\TwoAFCTireUGLo, \TwoAFCTireUGHi] & \TwoAFCSoapUG{} [\TwoAFCSoapUGLo, \TwoAFCSoapUGHi] \\
\bottomrule
\end{tabular}
\end{table}

\noindent\textbf{Forced-choice comparison with baselines.}
The rating studies cover ScoreDial alone, so we compare perceived progression with the baselines in a two-alternative forced-choice (2AFC) study on tire and soap, with the same recruitment and consent. Each screen shows two rows of five images generated from the same test seed at $c=0.1,\ldots,0.9$ from left to right: one row from ScoreDial and one from Numeric prompt, MPGD-flow, MPGD-flow$^{+}$, or UG-flow, giving 40 pairs over the four baselines and ten test seeds. Raters are told that both rows request an increasing level from left to right and choose the row that becomes more clearly slip-resistant (tire) or translucent (soap), judging the change rather than image quality; there is no tie option. ScoreDial's row is on top in a random half of the pairs for each rater, and 8 pairs are repeated with the rows swapped. Under criteria fixed before analysis, a rater is excluded for giving the same answer on every screen or for choosing the same method on a repeat and its original in at most 3 of the 8 repeats, leaving \TwoAFCTireRaters{} tire and \TwoAFCSoapRaters{} soap raters. Table~\ref{tab:2afc} reports the fraction of non-repeat judgments preferring ScoreDial. ScoreDial is preferred in \TwoAFCMinPct--\TwoAFCMaxPct\% of judgments across baselines and attributes.

\section{Conclusion}
\label{sec:concl}
ScoreDial couples learned target conditioning with error-adaptive feedback for requested-value control from a few hundred human ratings. One sampler configuration reaches every tested cell on all four attributes ($R$-MAE${}\le{}$\ControlMAEMax{}, $\rho=\ControlRhoAll$). Crucially, agreement with the guidance regressor $R$ is not equivalent to attribute control: MPGD-flow also reaches the requested values under $R$, while ScoreDial has larger independent-auditor displacement on every attribute, even against a stronger MPGD-flow setting, and the largest among all compared methods. On every attribute, feedback-only retains low $R$ error while auditor displacement drops by more than half, whereas conditioning-only lacks accurate requested-value agreement. Blinded human studies on tire and soap show mean ratings tracking $c$, and in forced-choice comparisons raters chose ScoreDial's progression as the clearer one in at least \TwoAFCMinPct\% of judgments against every baseline.

\noindent\textbf{Limitations.} The soap regressor is cross-validated by image rather than by physical soap, so its $R^2$ may be optimistic.

\FloatBarrier

\clearpage
\normalsize
\newcommand{\suppgrid}[3]{\begin{figure*}[p]\centering\includegraphics[width=\textwidth]{figs/#1}\caption{#2}\label{#3}\end{figure*}}
\setcounter{section}{0}\setcounter{figure}{0}\setcounter{table}{0}
\renewcommand{\thesection}{\Alph{section}}
\renewcommand{\thefigure}{S\arabic{figure}}
\renewcommand{\thetable}{S\arabic{table}}
\renewcommand{\theHsection}{supp.\Alph{section}}
\renewcommand{\theHfigure}{supp.\arabic{figure}}
\renewcommand{\theHtable}{supp.\arabic{table}}
\maketitlesupplementary

This supplement accompanies the main paper. Sections~\ref{sec:roundtrip}--\ref{sec:selection} give the details that the main paper refers to: the autoencoder round-trip check (Section~3.3 of the main paper), the training-pool memorization check (Section~4.1), and the regressor-selection protocol (Section~4.2). Sections~\ref{sec:all-comparison} and~\ref{sec:all-ablation} show every test image behind Tables~1 and~2.

\section{Autoencoder round-trip check}
\label{sec:roundtrip}
Section~3.3 of the main paper states that, on the rated images, the frozen autoencoder path alone changes $R$ by at most \RoundTripPctMax\% of the control range. Each of the \RoundTripImages{} rated images (118 tire, 100 interior, 479 food, 150 soap) is resized to $1024^2$, encoded and decoded by the frozen SD3.5 autoencoder used during sampling, and scored by the deployed $R$ before and after. Before running, we fixed the criterion that the round trip is negligible for an attribute if the mean $|\Delta R|$ is below 5\% of the control range (0.8, the width of the requested values $0.1$--$0.9$) and the Pearson correlation between the two scores is at least 0.99. The mean $|\Delta R|$ is \RoundTripTirePct\%, \RoundTripInteriorPct\%, \RoundTripFoodPct\%, and \RoundTripSoapPct\% of the range for tire, interior, food, and soap, with $r\geq\RoundTripRMin$, so every attribute meets the criterion. As a reference, pairing each original with a randomly permuted reconstruction gives the $|\Delta R|$ between unrelated images of the same attribute; the actual round-trip shift is at most \RoundTripShuffledMax{} of that value. The check isolates the decoder path; attribute movement is evaluated separately in Section~4.3 of the main paper.

\section{Training-pool memorization check}
\label{sec:memorization}
Section~4.1 of the main paper states that an SSCD copy-detection check finds no reported image unusually close to its generator-training pool. For each attribute we take the 50 reported ScoreDial images (the Full cells of Tables~1 and~2) and the generator-training pool of Section~4.1, embed both with the released SSCD descriptor \texttt{sscd\_disc\_mixup} (Pizzi et al., CVPR 2022), and measure cosine similarity. Each generated image is matched to its nearest training image. As a reference, each training image's nearest neighbour within the pool, excluding itself, gives the similarity the pool already reaches internally. Table~\ref{tab:sscd} lists the results: on every attribute, even the closest generated image lies inside the pool's own nearest-neighbour distribution, at its 15th--74th percentile. The tire pool contains flipped copies of its tail images (Section~4.1), which raises its internal similarity.

\begin{table}[t]
\centering\footnotesize
\setlength{\tabcolsep}{3pt}
\caption{SSCD nearest-neighbour cosine similarity between the 50 reported images per attribute and the generator-training pool. ``Pctl.'' is the percentile of the largest generated-image similarity within the pool's internal nearest-neighbour distribution.}
\label{tab:sscd}
\begin{tabular}{lccccc}
\toprule
& Pool & \multicolumn{2}{c}{Generated NN} & Pool NN & \\
\cmidrule(lr){3-4}
Attribute & size & mean & max & mean & Pctl. \\
\midrule
Tire & 984 & 0.59 & 0.69 & 0.87 & 15.5 \\
Interior & 100 & 0.26 & 0.32 & 0.28 & 74.0 \\
Food & 479 & 0.50 & 0.66 & 0.66 & 48.2 \\
Soap & 800 & 0.44 & 0.57 & 0.56 & 57.9 \\
\bottomrule
\end{tabular}
\end{table}

\section{Regressor-selection protocol}
\label{sec:selection}
Section~4.2 of the main paper reports repeated nested cross-validation of the full selection procedure. In each of five repeats, the labelled images are split into five outer folds. Within each outer-training split, a five-fold inner cross-validation scores all 28 configurations, the seven backbone--representation candidates of Section~3.1 combined with ridge $\alpha\in\{1,10,50,100\}$, and keeps the one with the highest inner $R^2$. That configuration is refit on the outer-training split and predicts the held-out outer fold. Each repeat yields one $R^2$ over all outer predictions, and the main paper reports their mean $\pm$ SD. These scores estimate the selection procedure as a whole: the configurations chosen in the outer folds are recorded only to inspect their stability and do not choose the deployed regressor, which is selected separately as described in Section~4.2.

\clearpage
\section{All test images of the matched comparison}
\label{sec:all-comparison}
Table~1 of the main paper is computed from \ExtTireOursN{} test images per method and attribute (five requested values $\times$ ten test seeds), and Figure~1 shows one seed per attribute. Figures~\ref{fig:cmp-tire-1}--\ref{fig:cmp-soap-3} show all of them. Each block is one test seed, with the methods as rows and the requested values $c=0.1,0.3,\ldots,0.9$ as columns; within a block, all methods use the same seed. MPGD-flow is shown at its selected strength and MPGD-flow$^{+}$ one strength rung above it.

\suppgrid{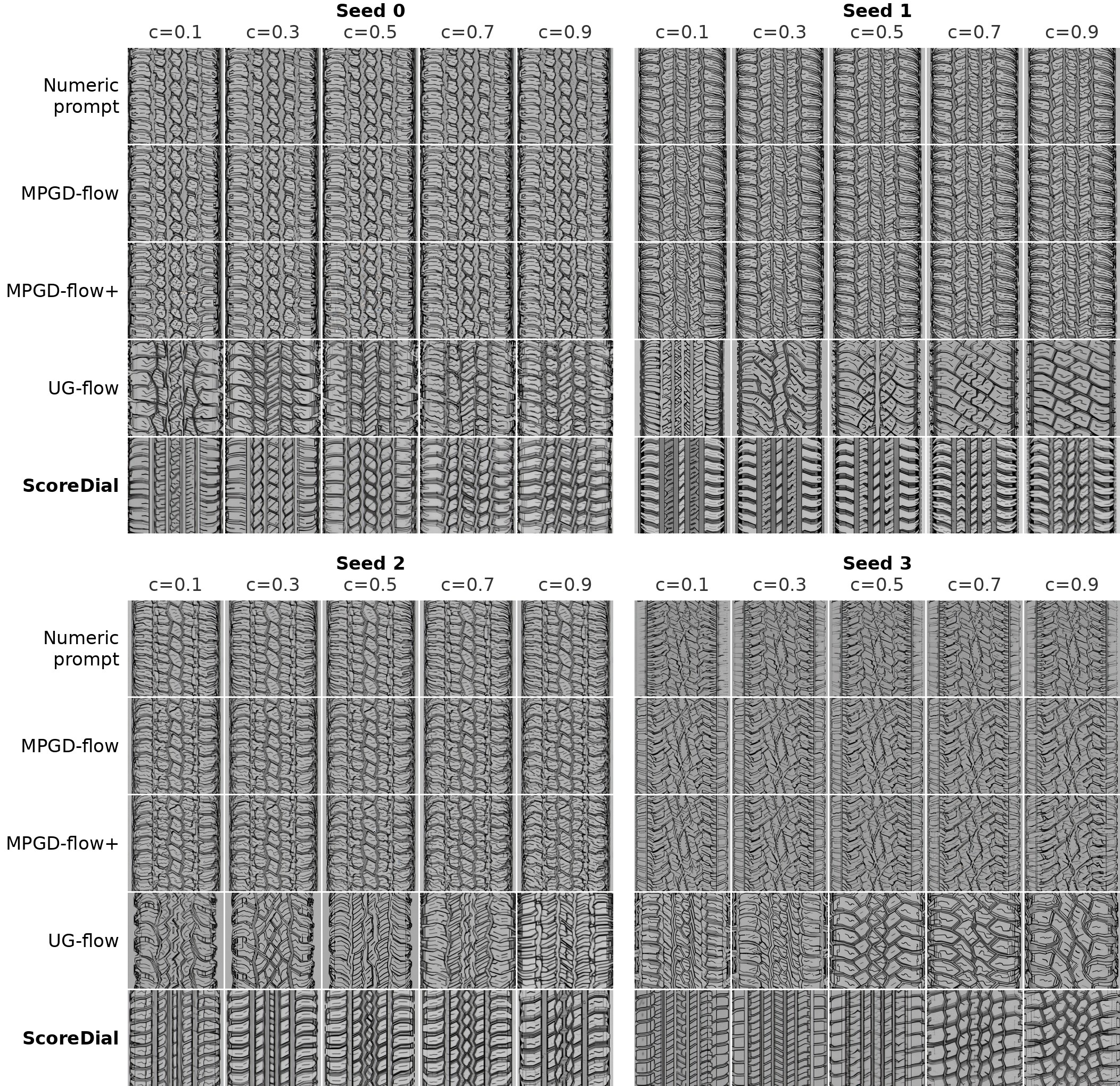}{Tire slip resistance, matched comparison, test seeds 0--3.}{fig:cmp-tire-1}
\suppgrid{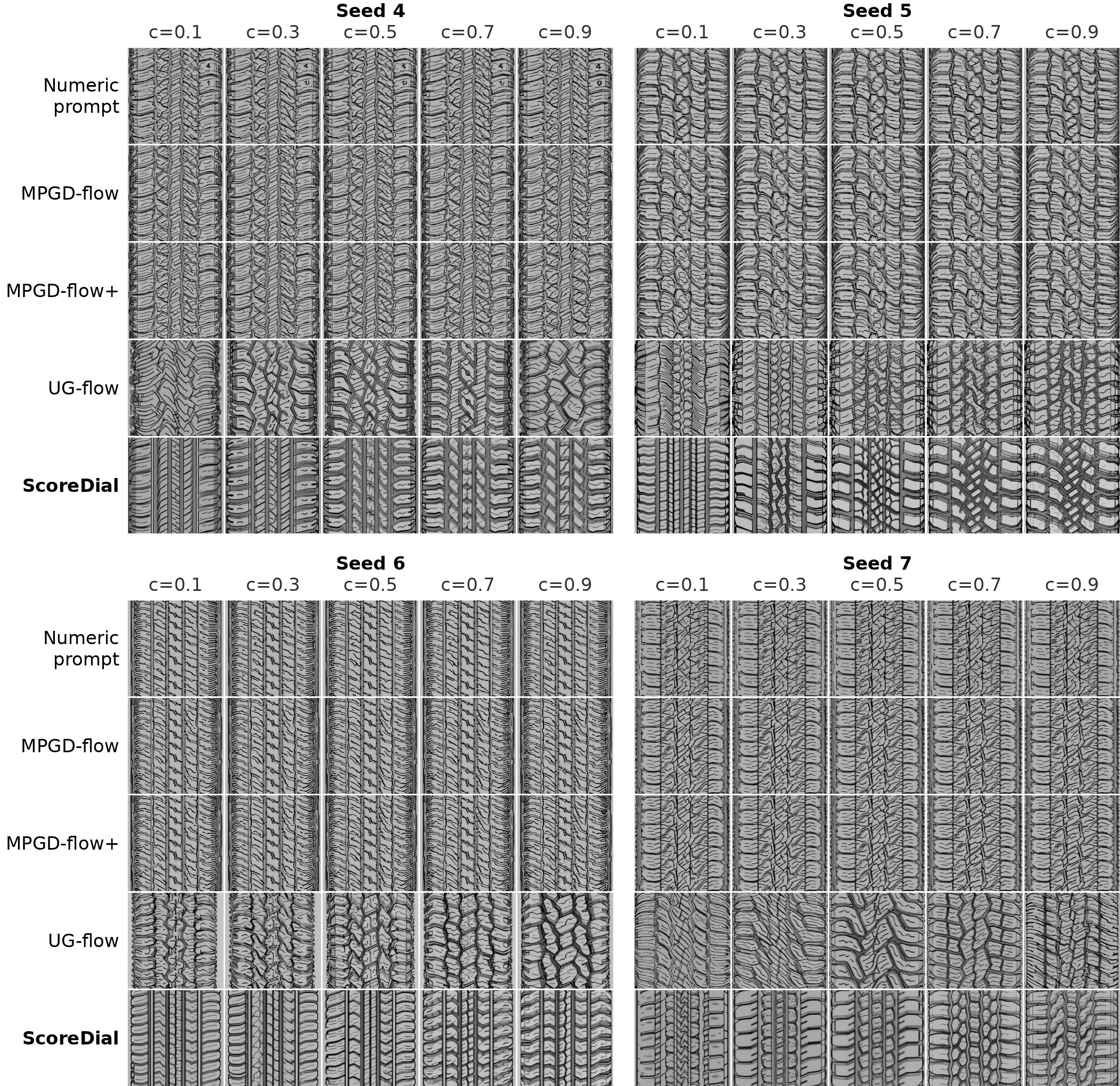}{Tire slip resistance, matched comparison, test seeds 4--7.}{fig:cmp-tire-2}
\suppgrid{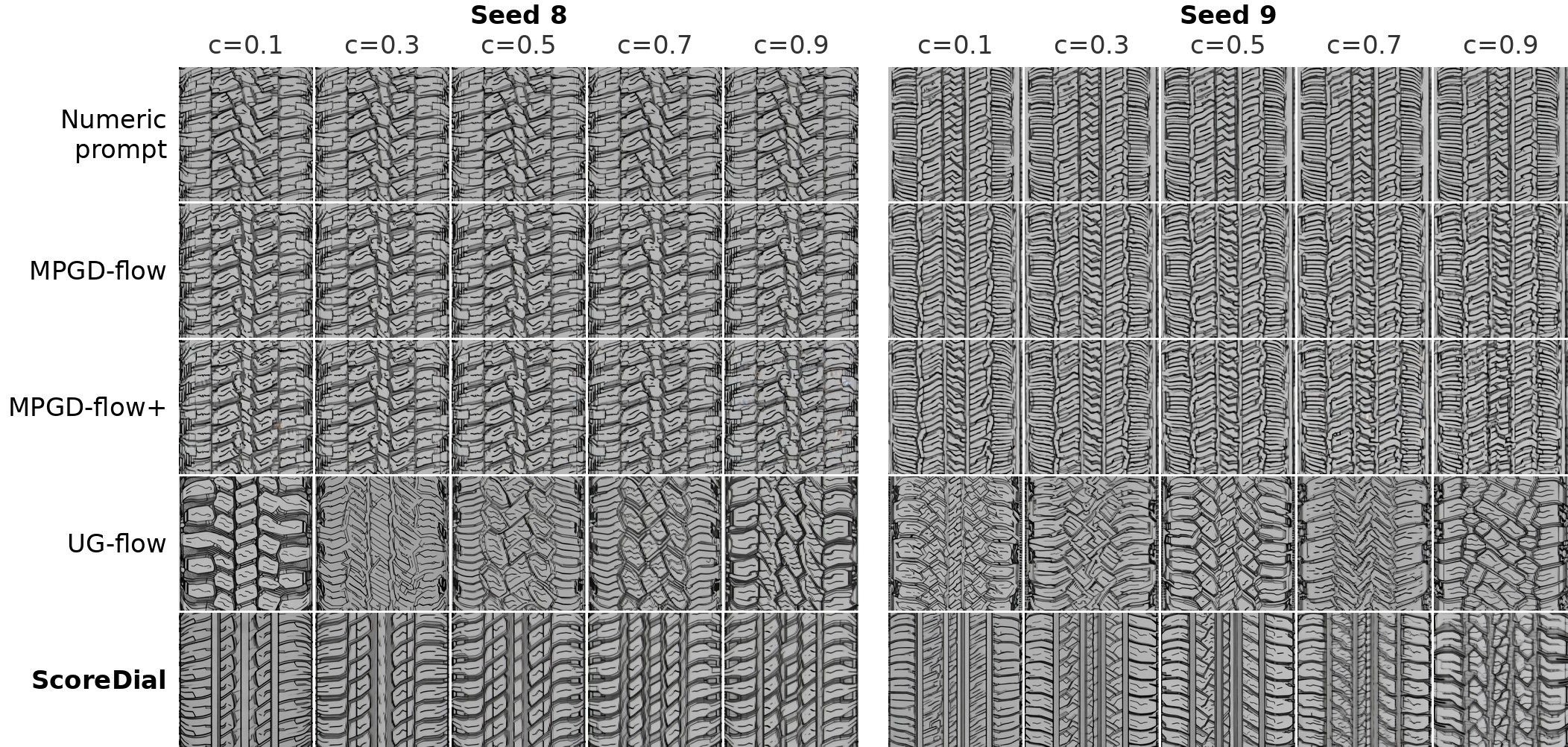}{Tire slip resistance, matched comparison, test seeds 8--9.}{fig:cmp-tire-3}
\suppgrid{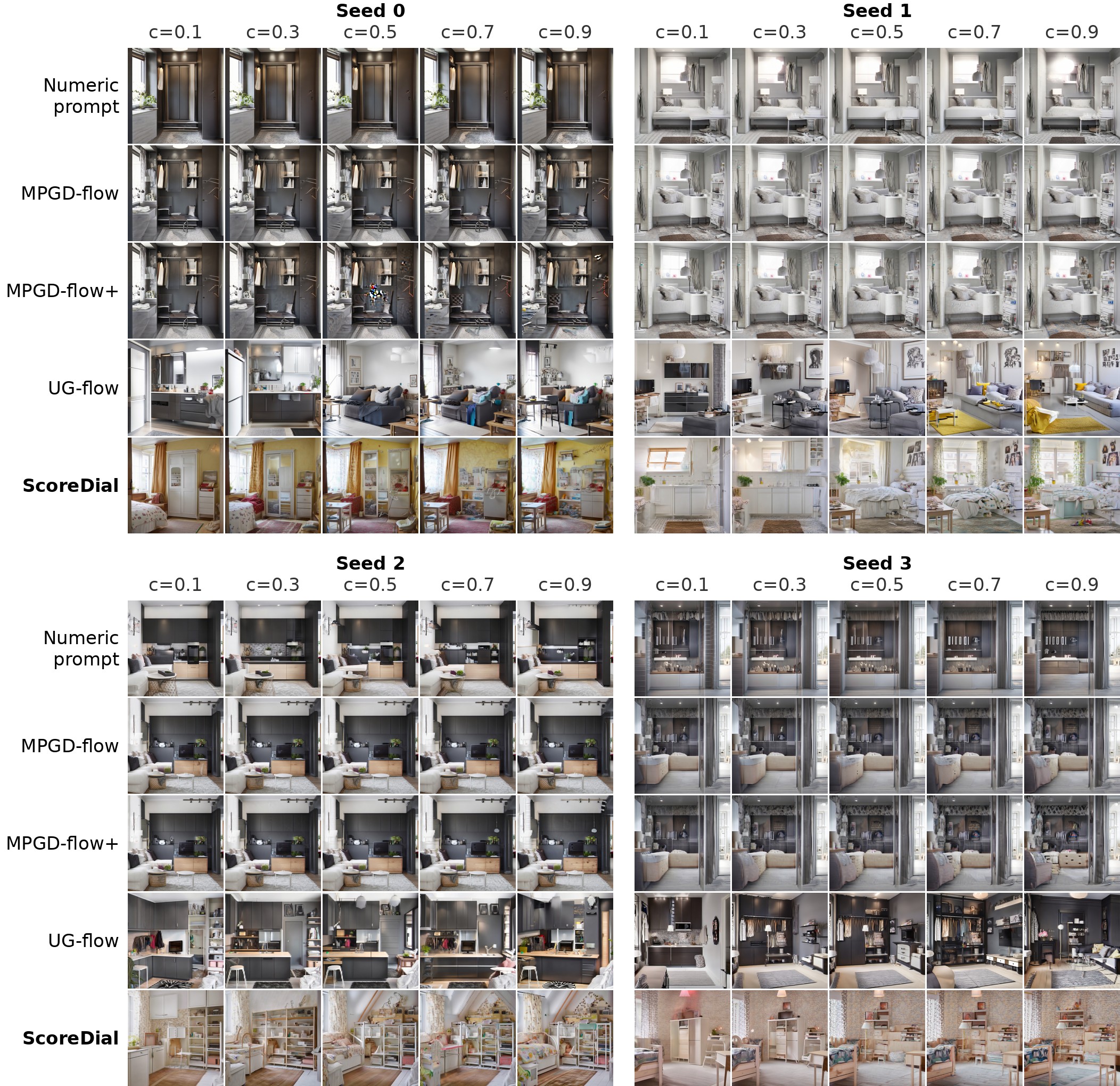}{Interior visual complexity, matched comparison, test seeds 0--3.}{fig:cmp-interior-1}
\suppgrid{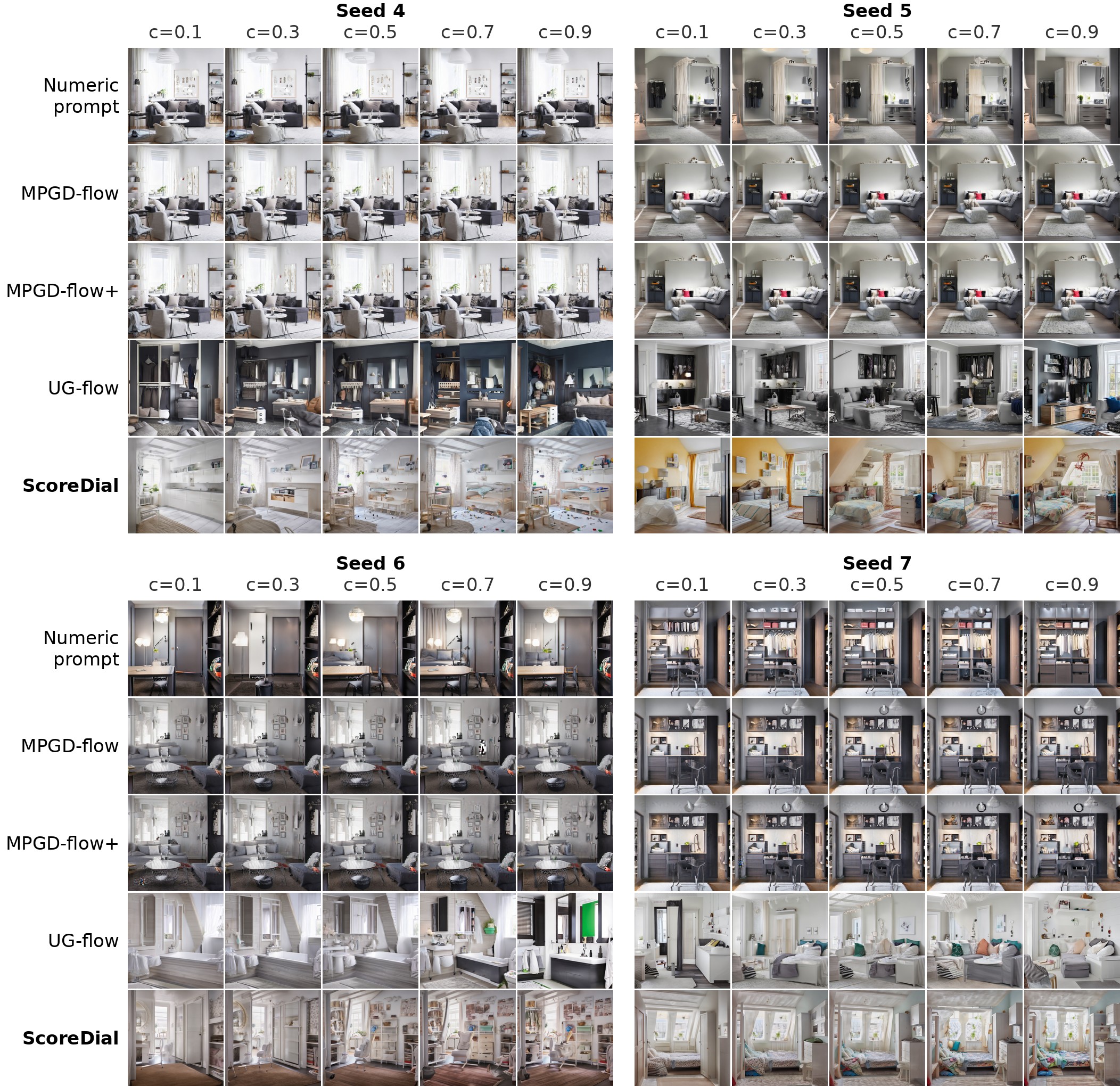}{Interior visual complexity, matched comparison, test seeds 4--7.}{fig:cmp-interior-2}
\suppgrid{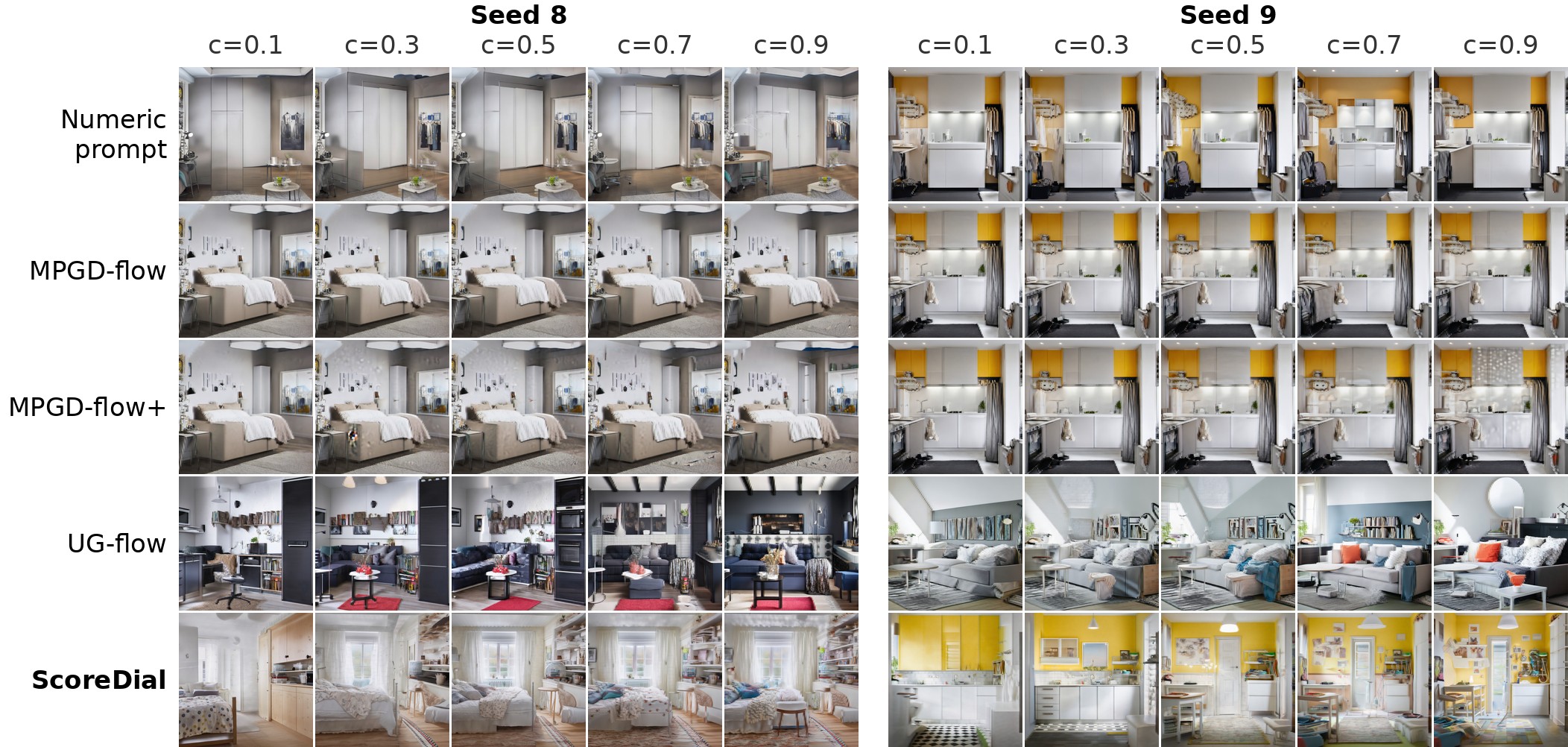}{Interior visual complexity, matched comparison, test seeds 8--9.}{fig:cmp-interior-3}
\suppgrid{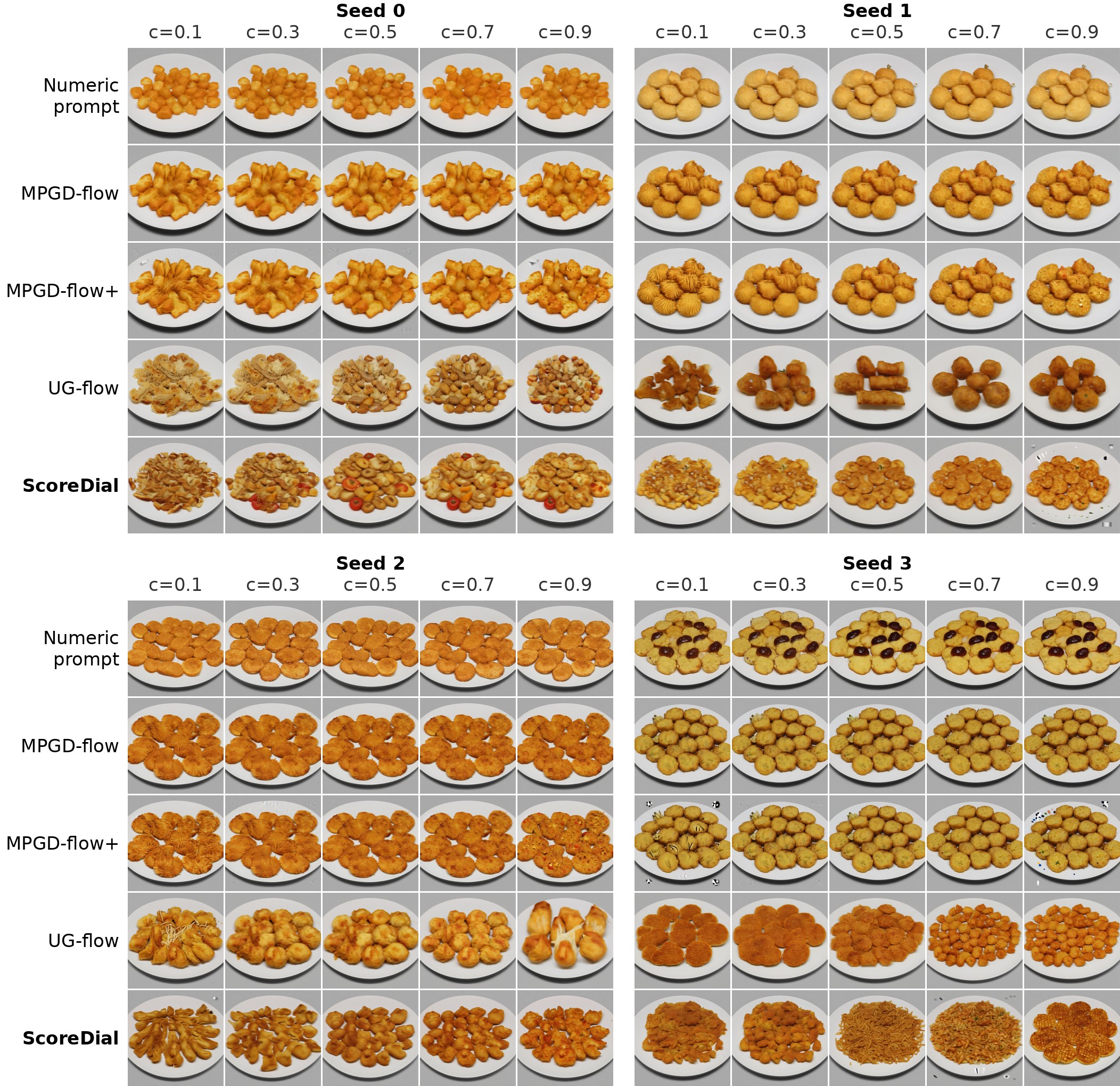}{Food desire-to-eat, matched comparison, test seeds 0--3.}{fig:cmp-food-1}
\suppgrid{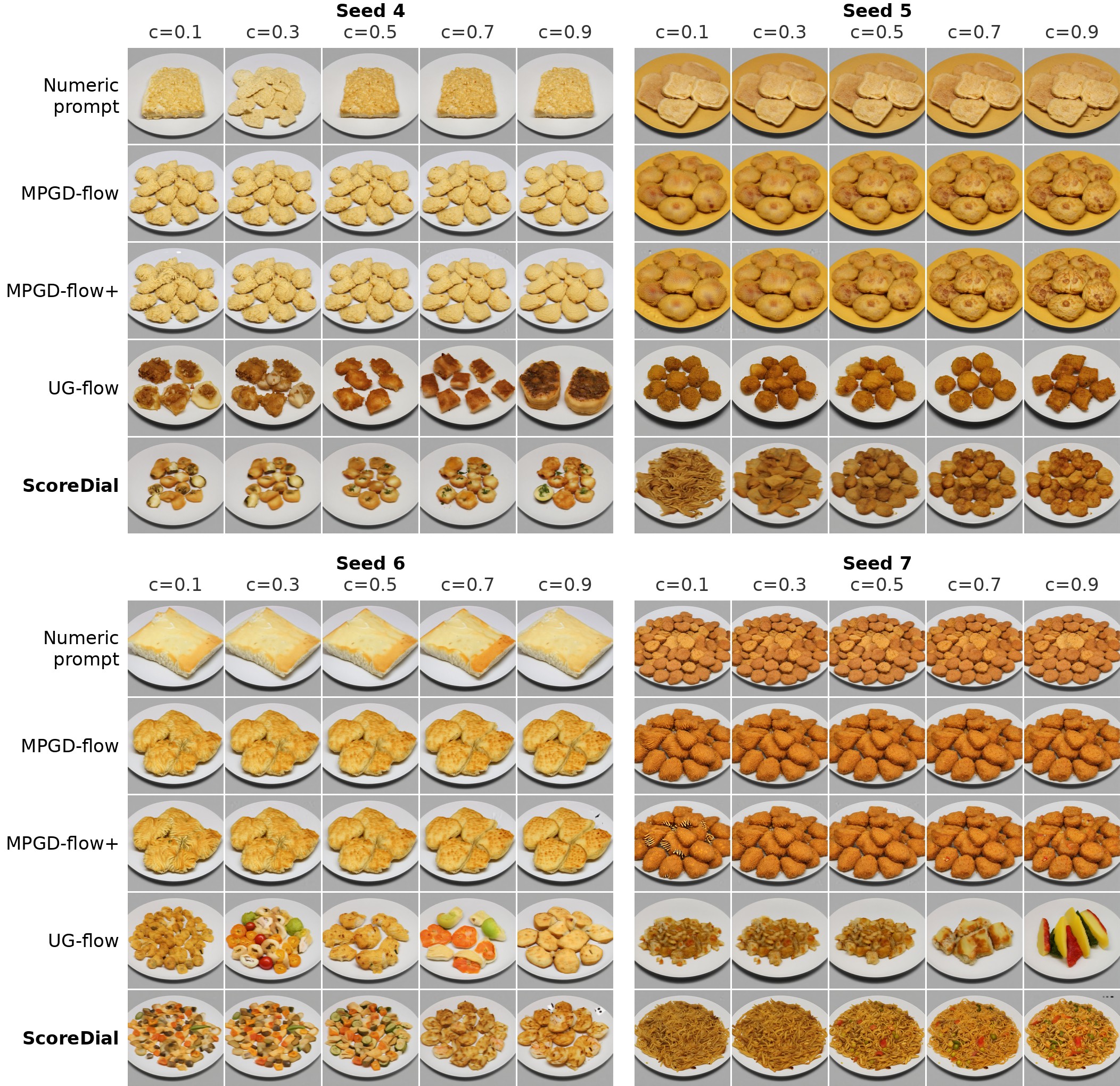}{Food desire-to-eat, matched comparison, test seeds 4--7.}{fig:cmp-food-2}
\suppgrid{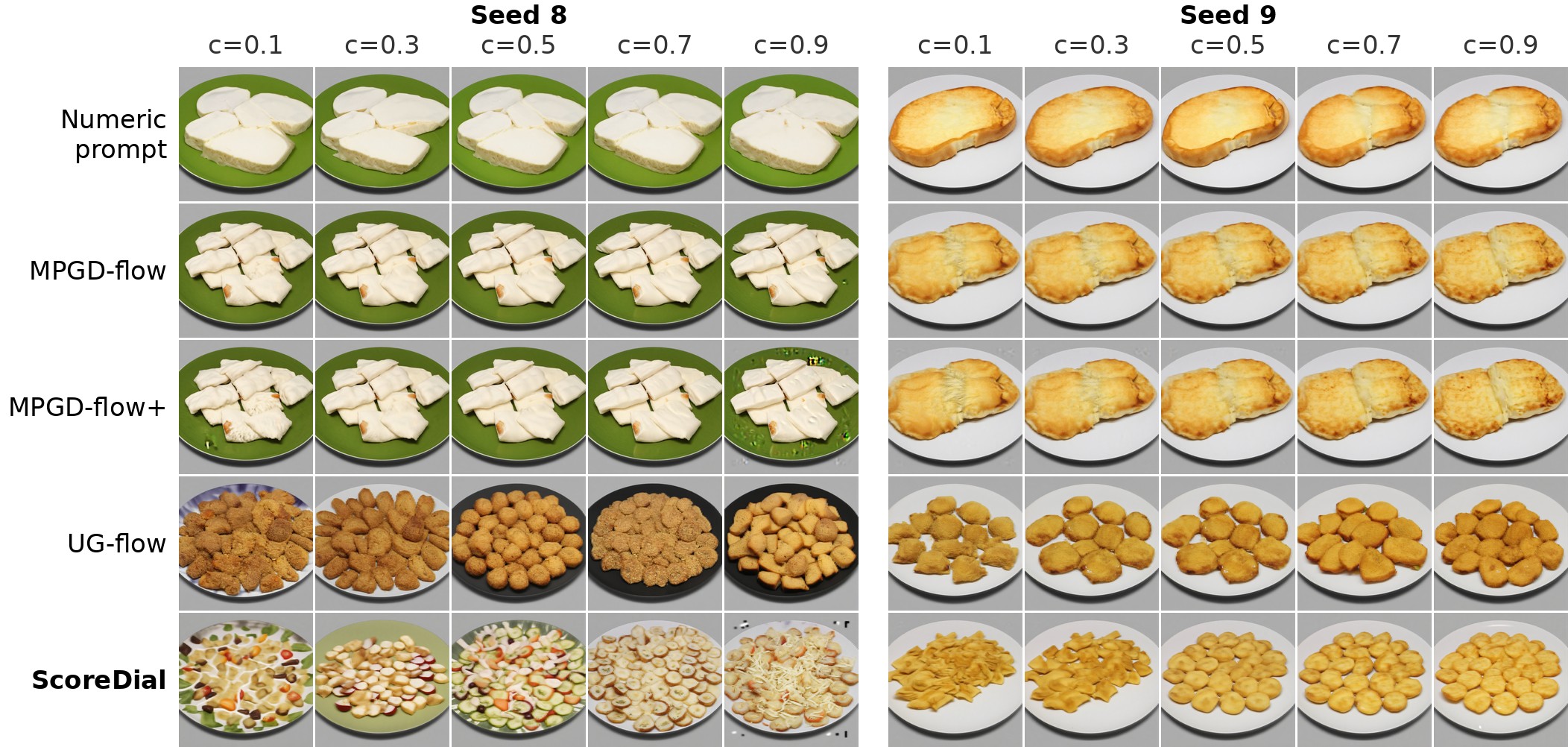}{Food desire-to-eat, matched comparison, test seeds 8--9.}{fig:cmp-food-3}
\suppgrid{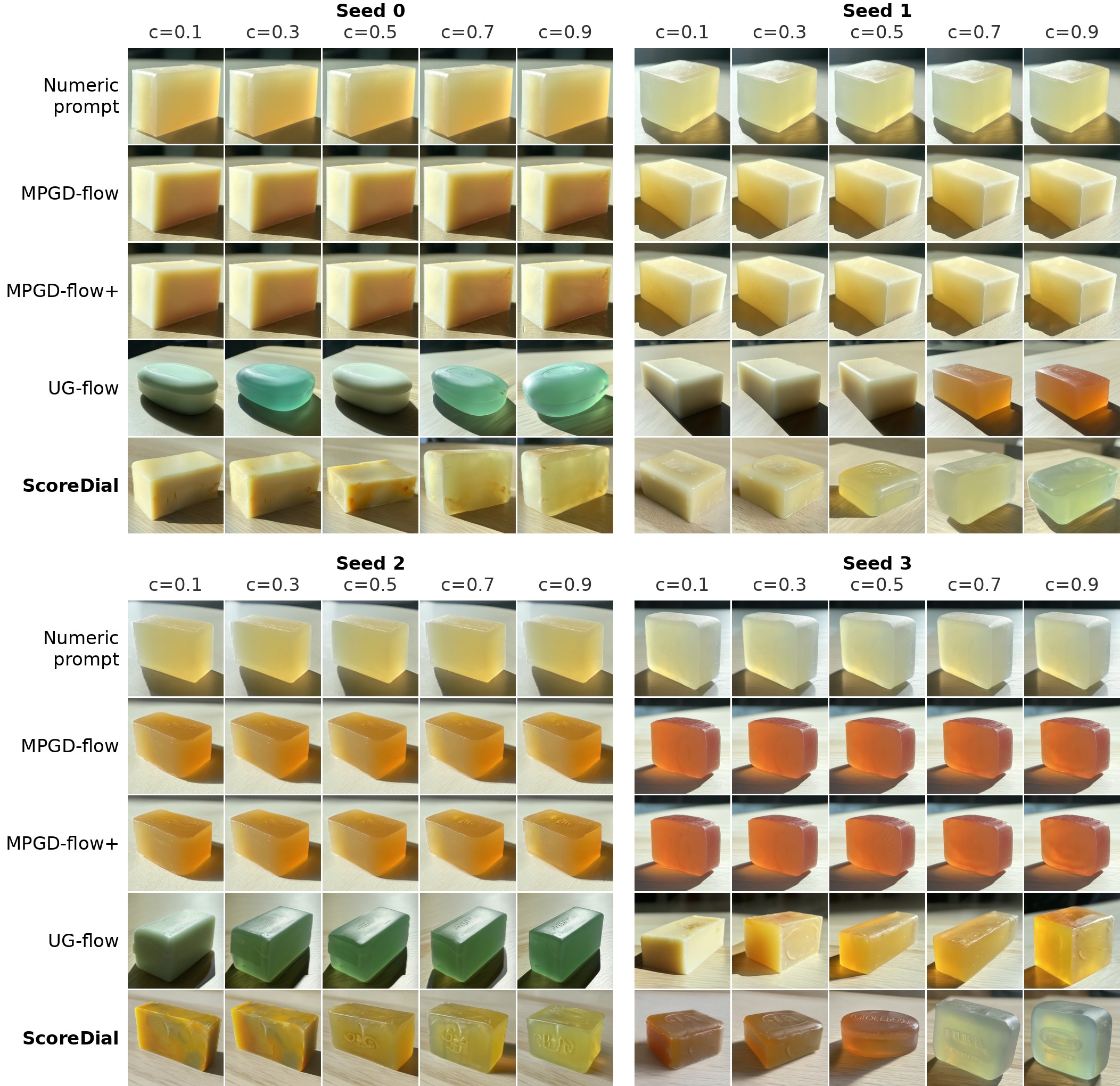}{Soap translucency, matched comparison, test seeds 0--3.}{fig:cmp-soap-1}
\suppgrid{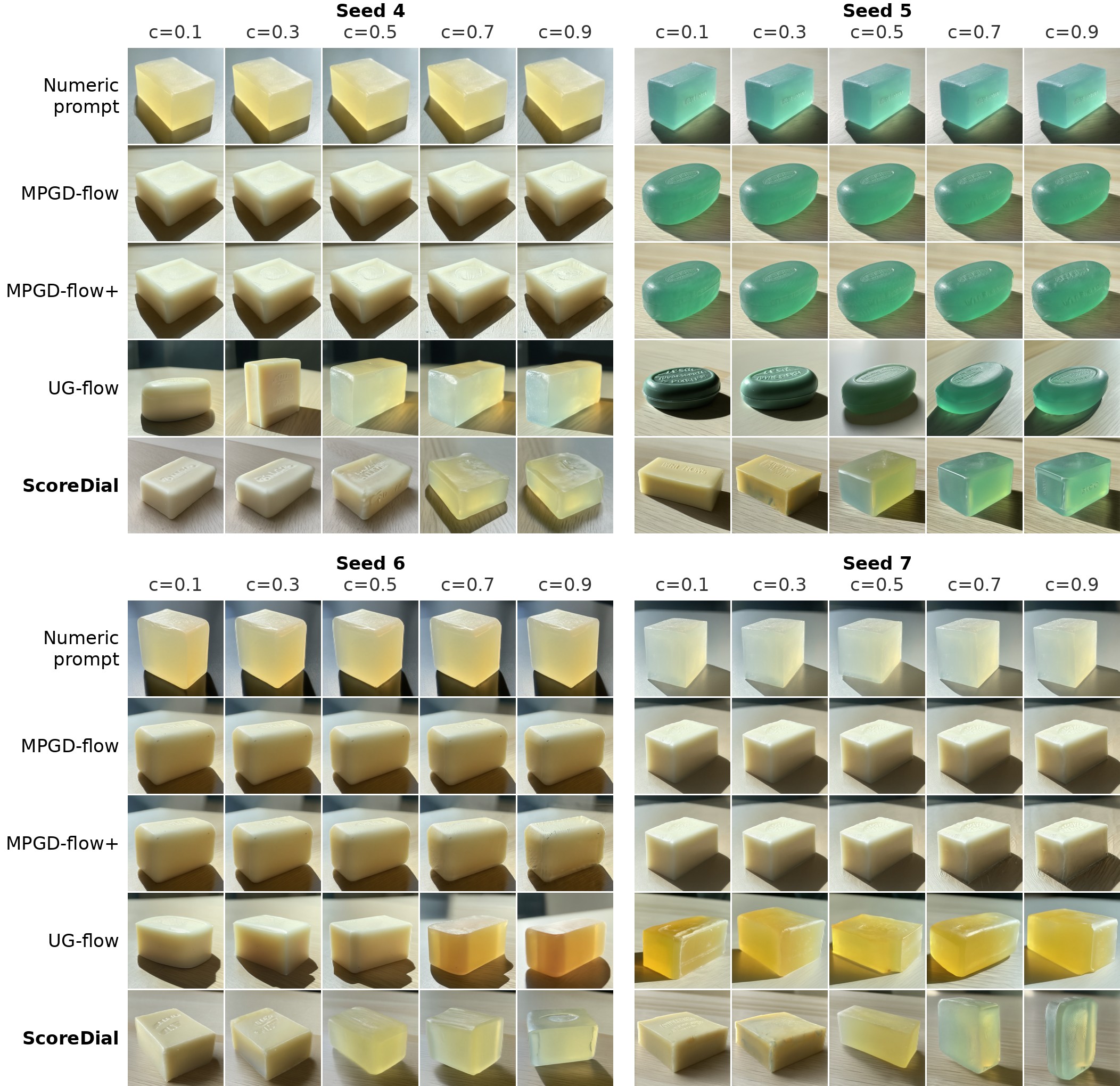}{Soap translucency, matched comparison, test seeds 4--7.}{fig:cmp-soap-2}
\suppgrid{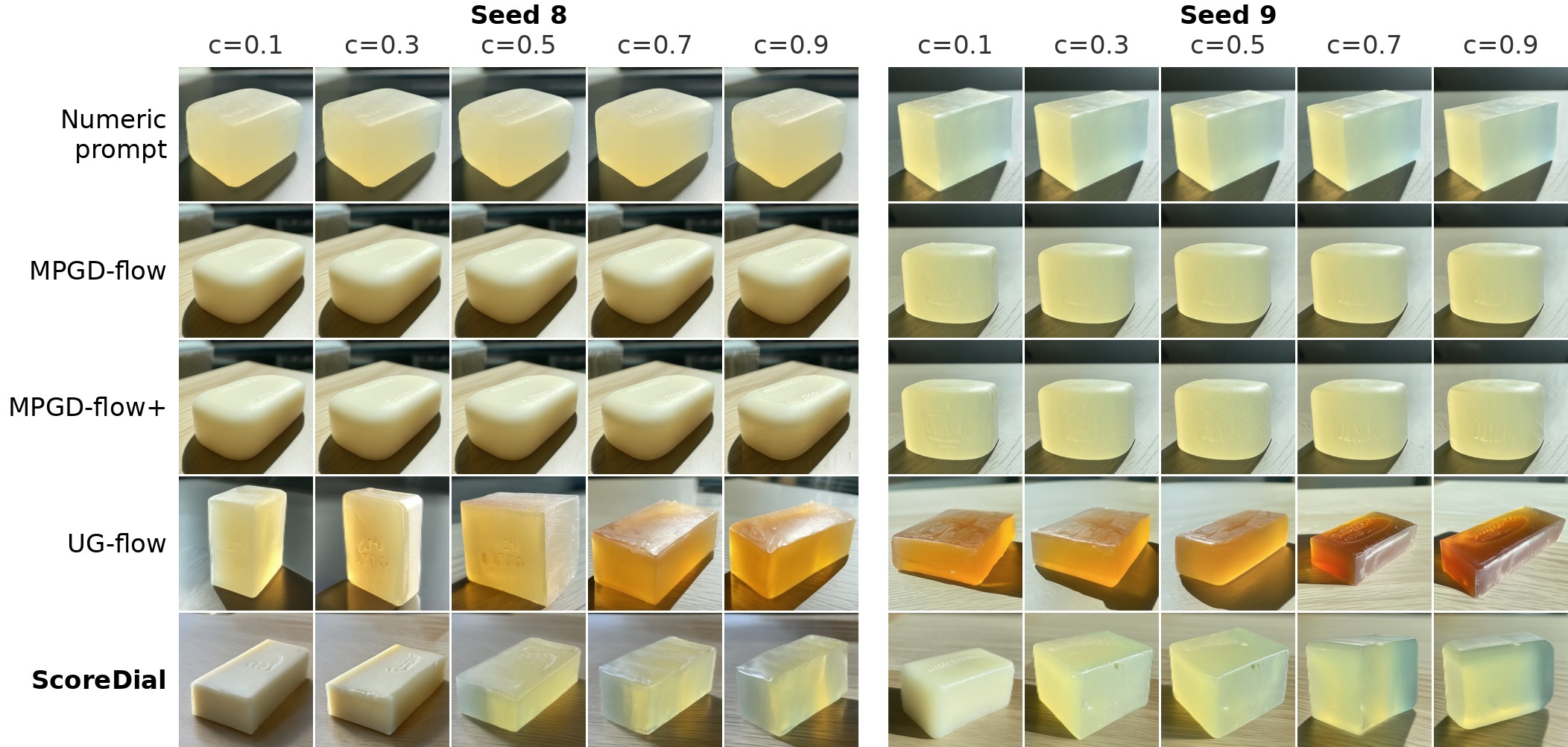}{Soap translucency, matched comparison, test seeds 8--9.}{fig:cmp-soap-3}

\clearpage
\section{All test images of the ablation}
\label{sec:all-ablation}
Table~2 of the main paper reports five configurations on the same $5\times10$ target--seed grid, and Figure~3 shows Full and G-only for tire at three requested values. Figures~\ref{fig:abl-tire-1}--\ref{fig:abl-soap-3} show every image of every configuration and attribute in the layout of Section~\ref{sec:all-comparison}, with the five configurations as rows. Full is the ScoreDial run of Table~1, so its row repeats the ScoreDial row of the corresponding comparison figure.

\suppgrid{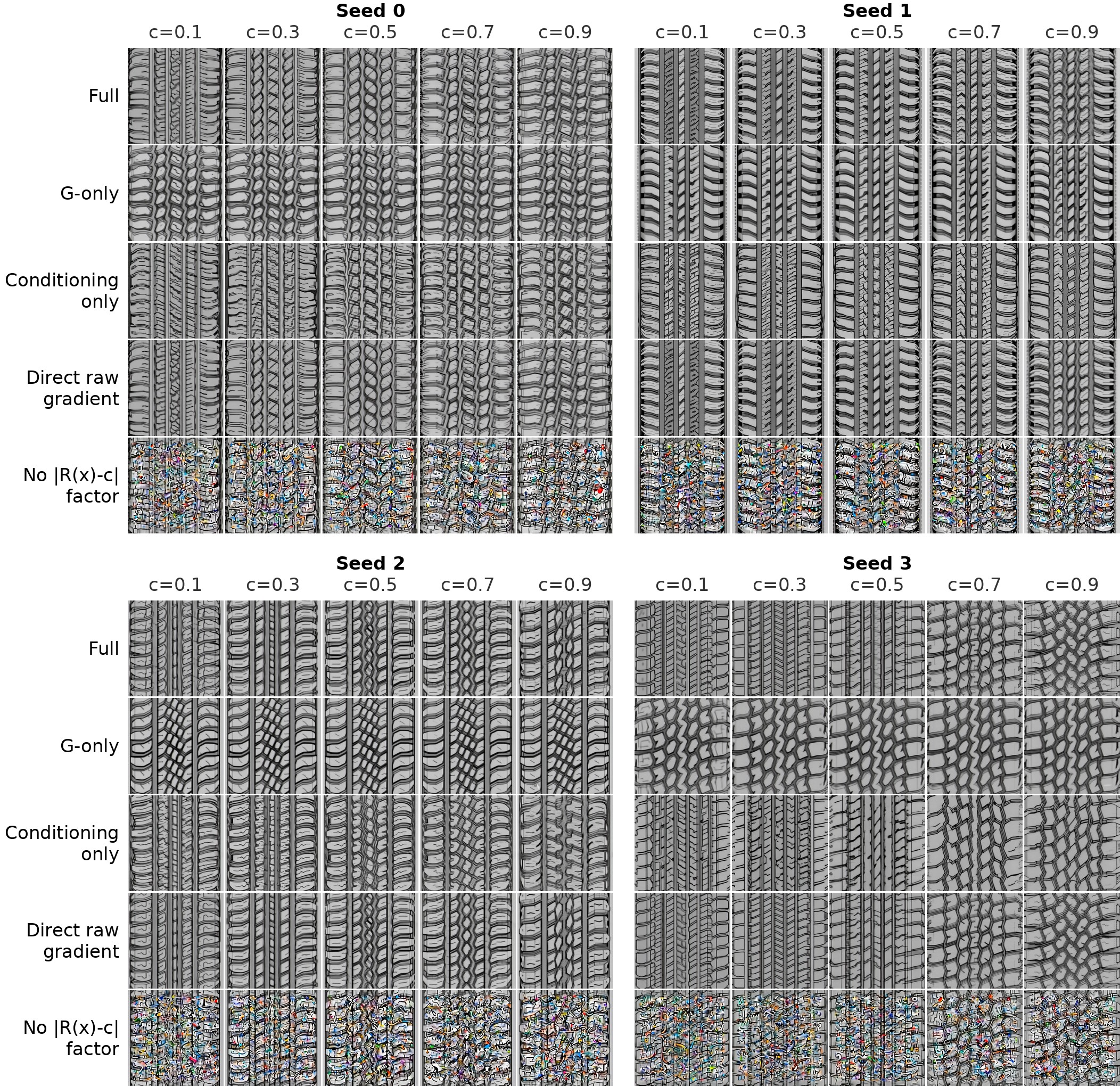}{Tire slip resistance, ablation, test seeds 0--3.}{fig:abl-tire-1}
\suppgrid{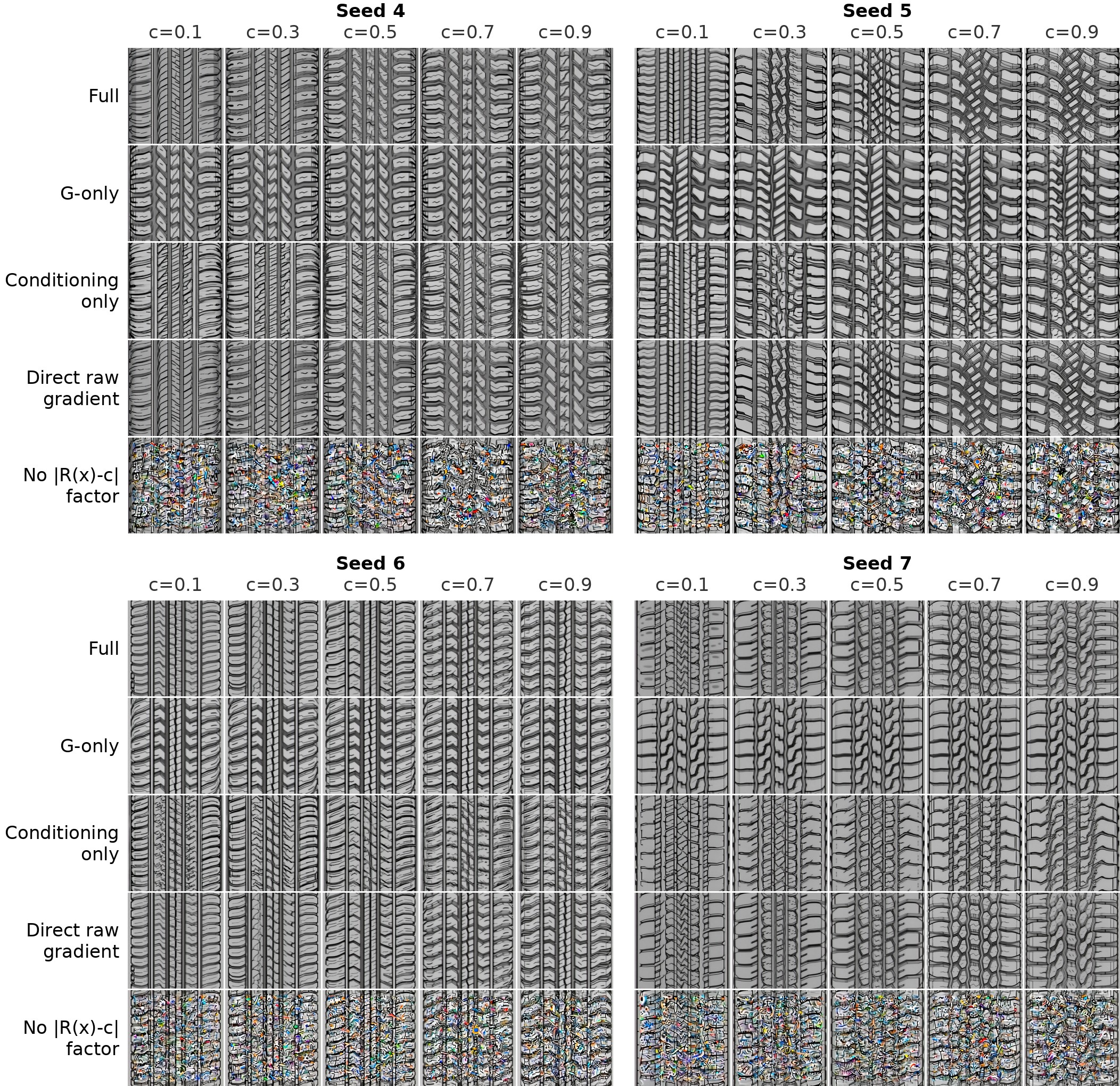}{Tire slip resistance, ablation, test seeds 4--7.}{fig:abl-tire-2}
\suppgrid{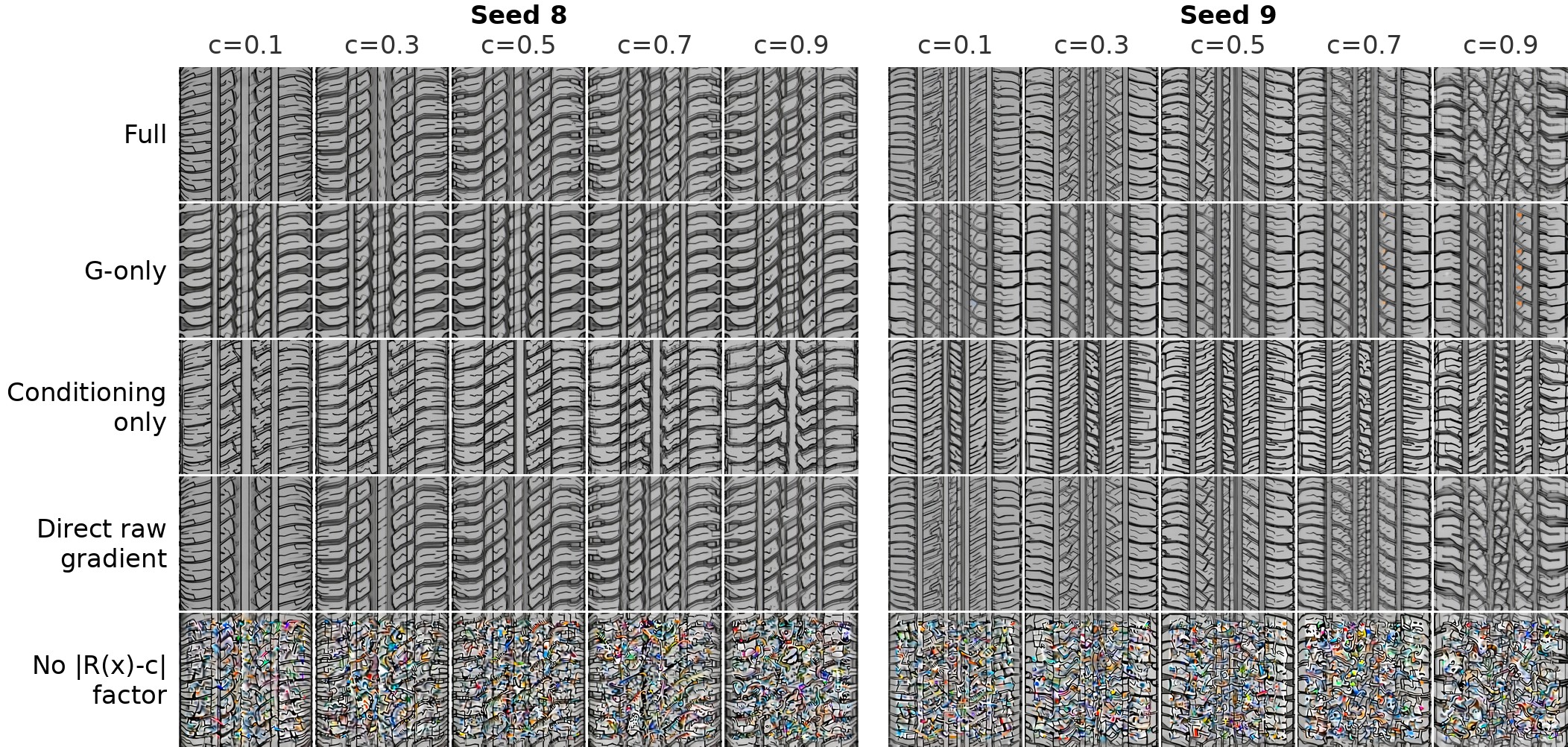}{Tire slip resistance, ablation, test seeds 8--9.}{fig:abl-tire-3}
\suppgrid{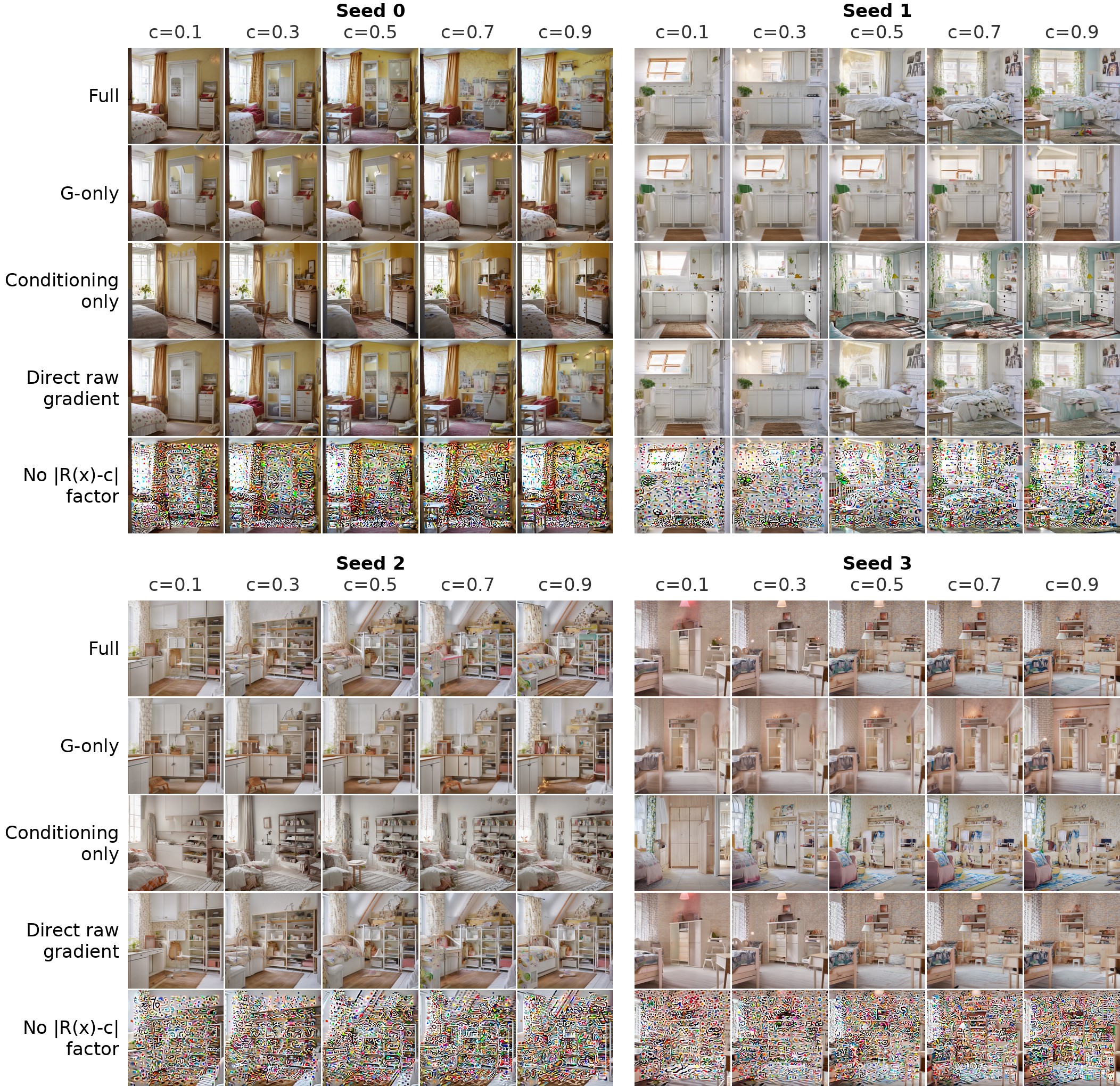}{Interior visual complexity, ablation, test seeds 0--3.}{fig:abl-interior-1}
\suppgrid{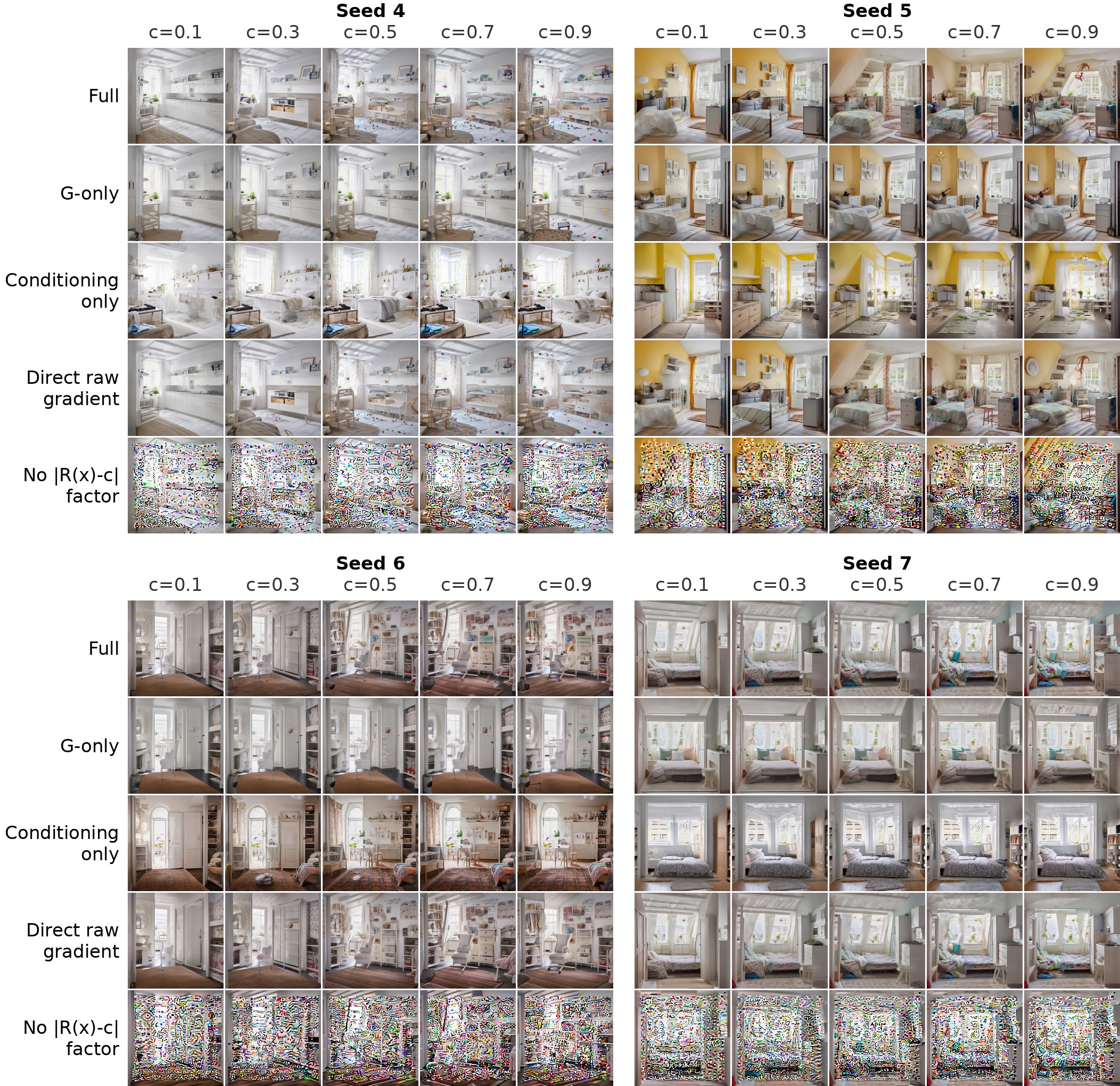}{Interior visual complexity, ablation, test seeds 4--7.}{fig:abl-interior-2}
\suppgrid{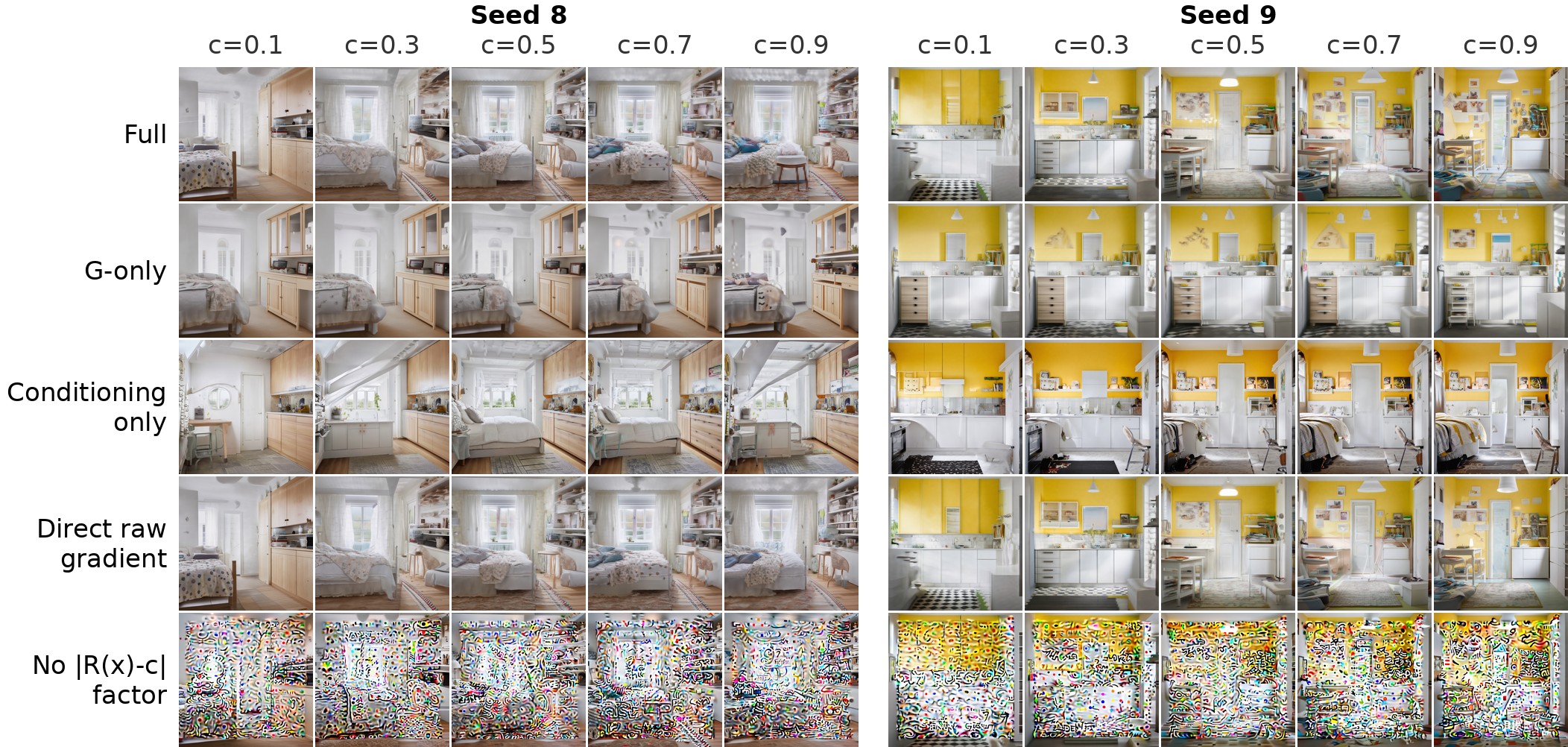}{Interior visual complexity, ablation, test seeds 8--9.}{fig:abl-interior-3}
\suppgrid{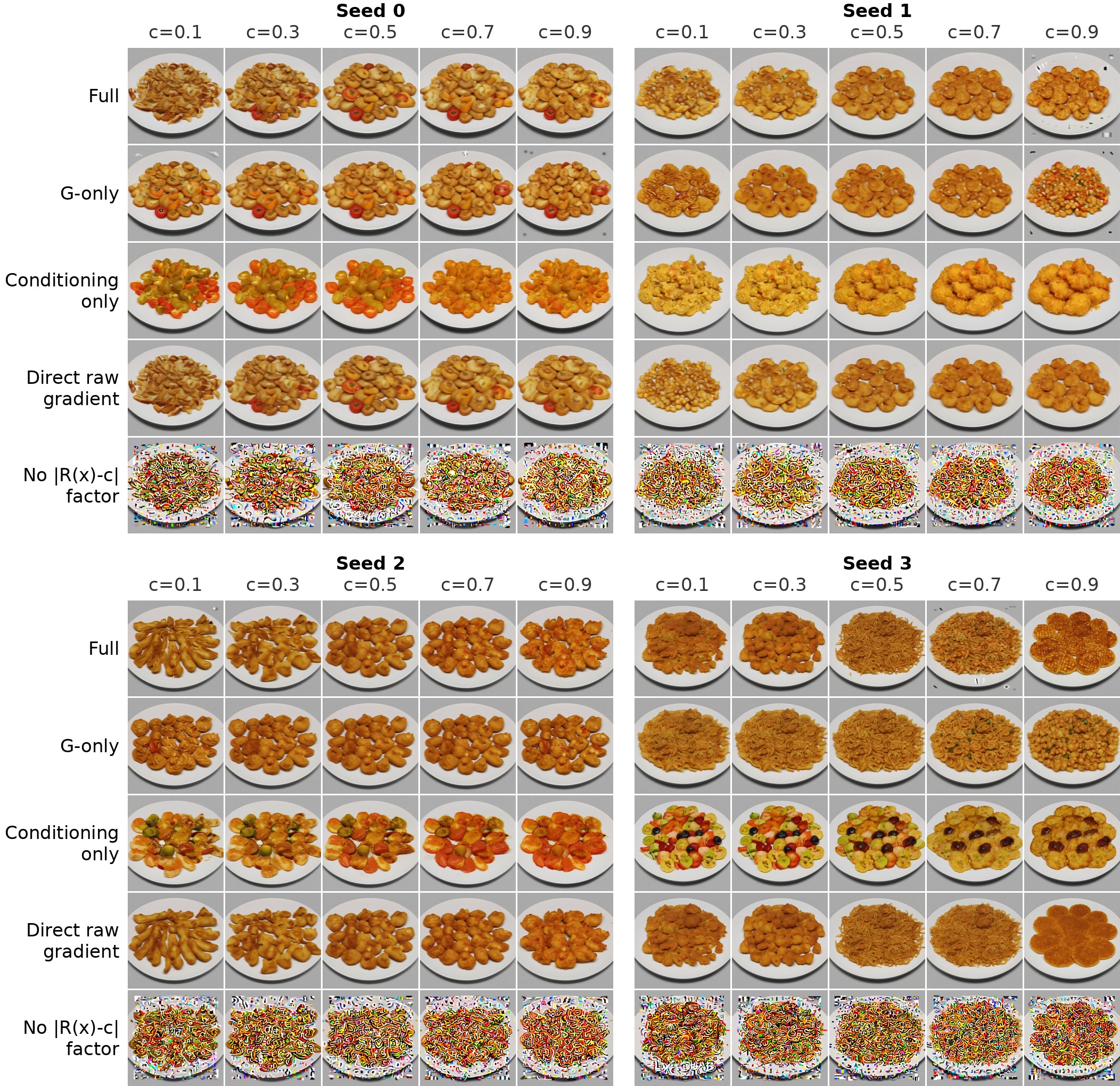}{Food desire-to-eat, ablation, test seeds 0--3.}{fig:abl-food-1}
\suppgrid{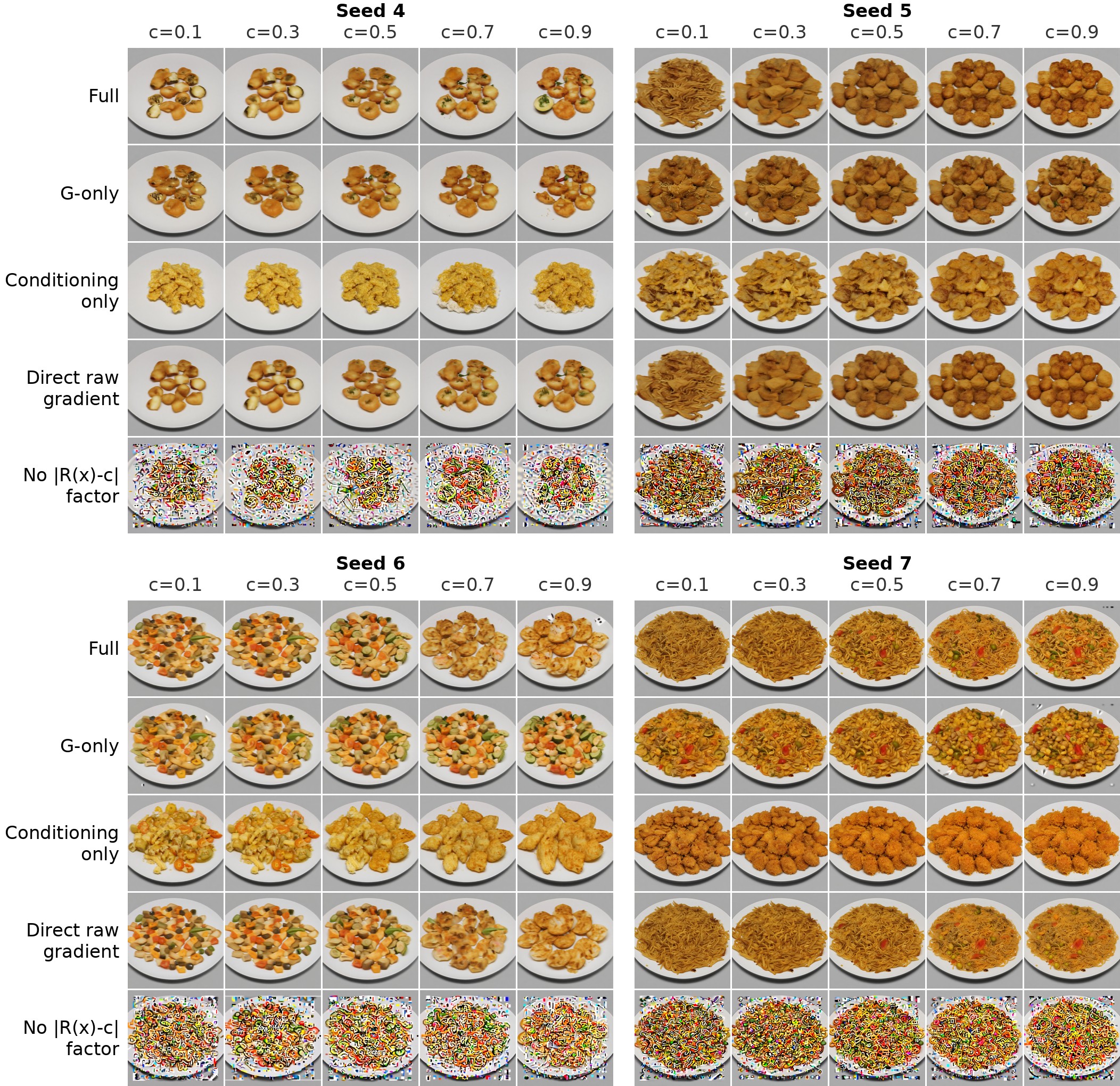}{Food desire-to-eat, ablation, test seeds 4--7.}{fig:abl-food-2}
\suppgrid{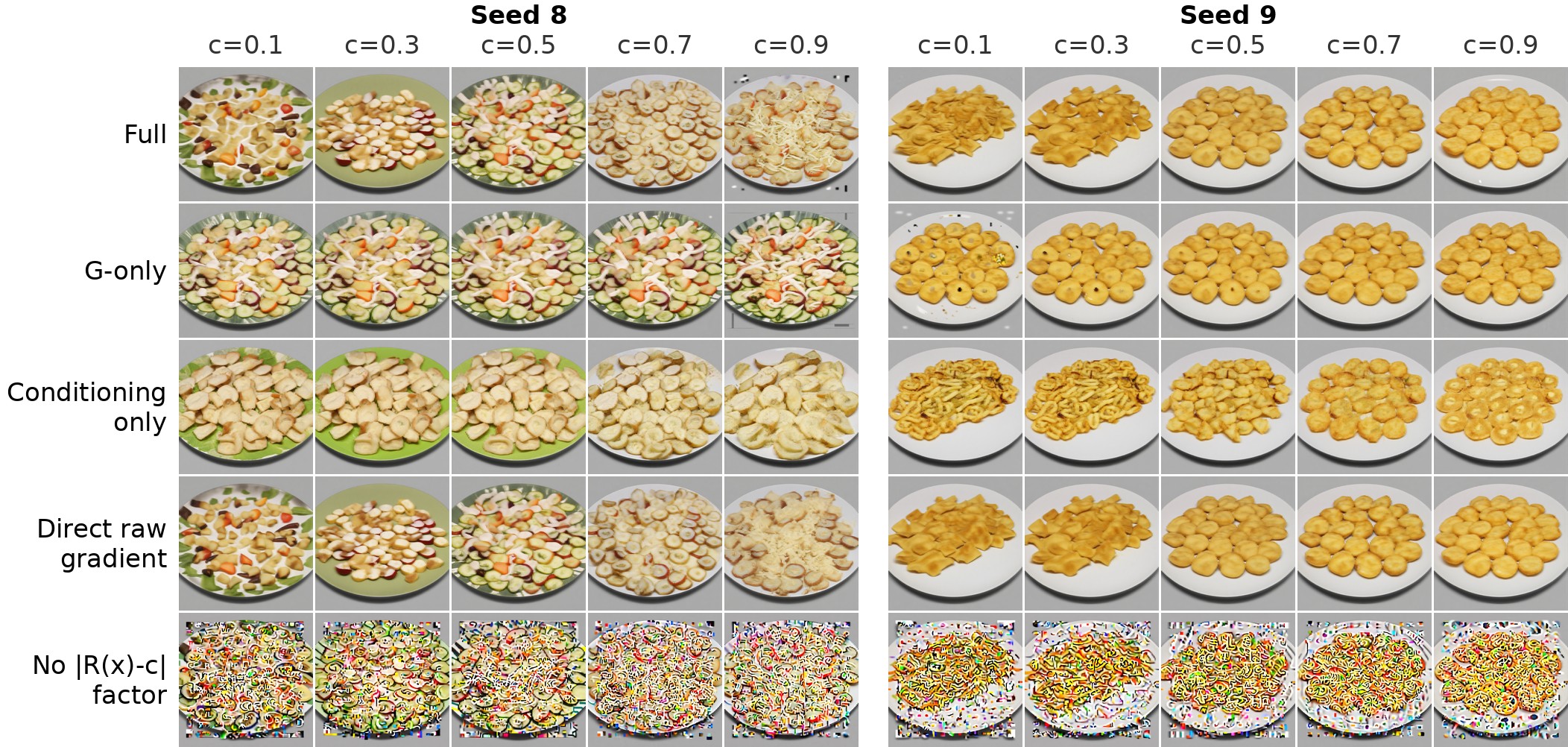}{Food desire-to-eat, ablation, test seeds 8--9.}{fig:abl-food-3}
\suppgrid{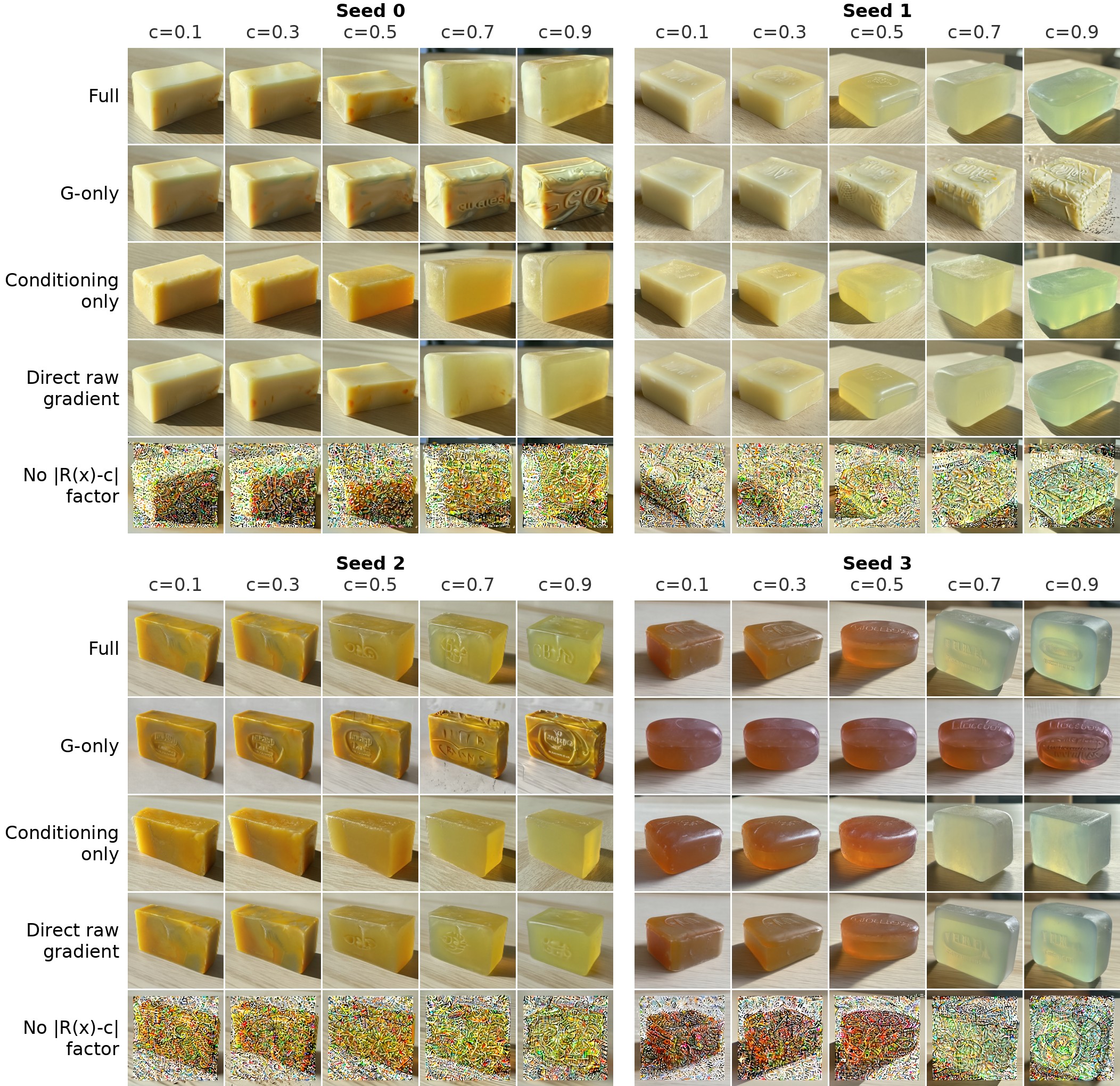}{Soap translucency, ablation, test seeds 0--3.}{fig:abl-soap-1}
\suppgrid{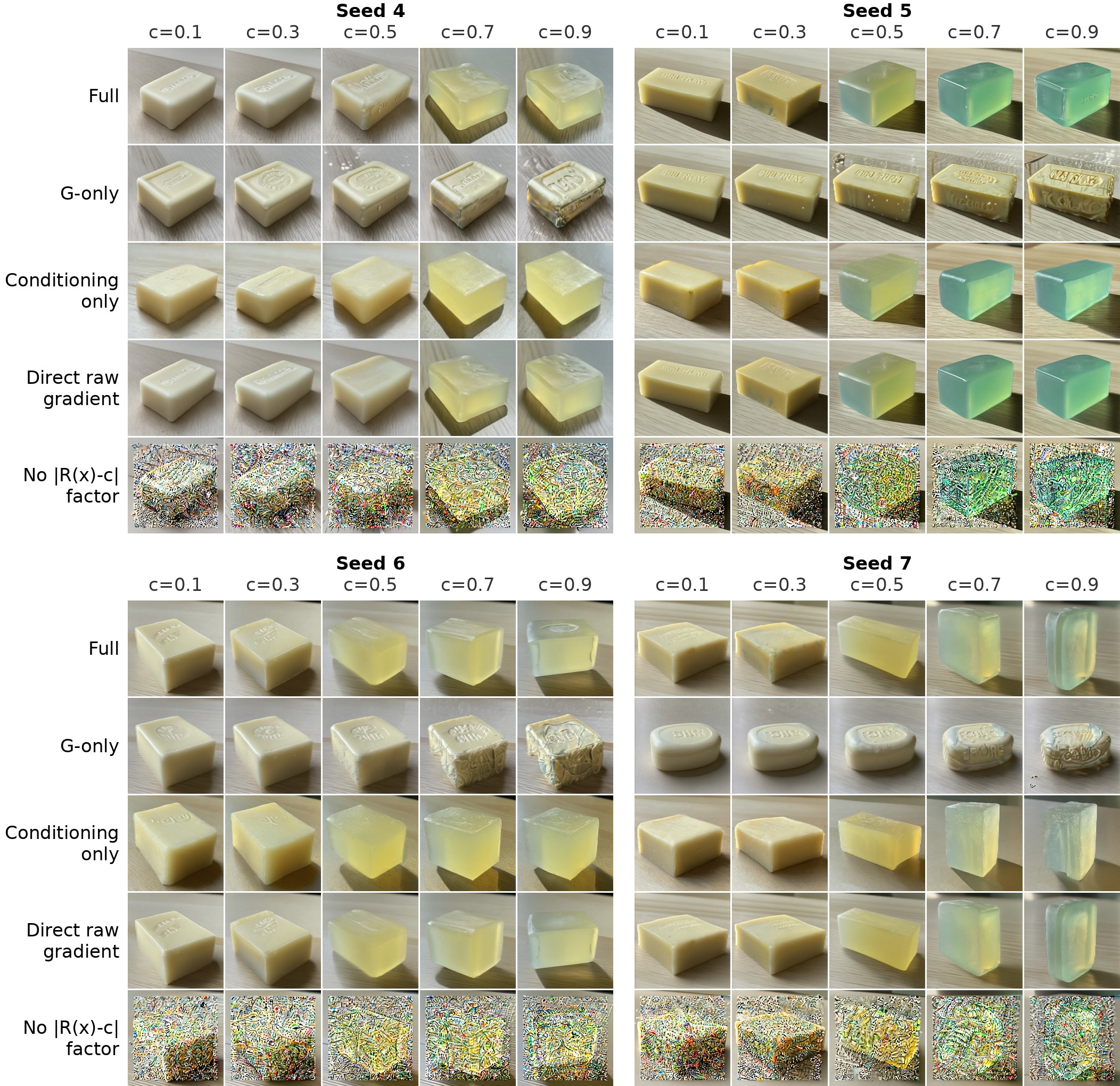}{Soap translucency, ablation, test seeds 4--7.}{fig:abl-soap-2}
\suppgrid{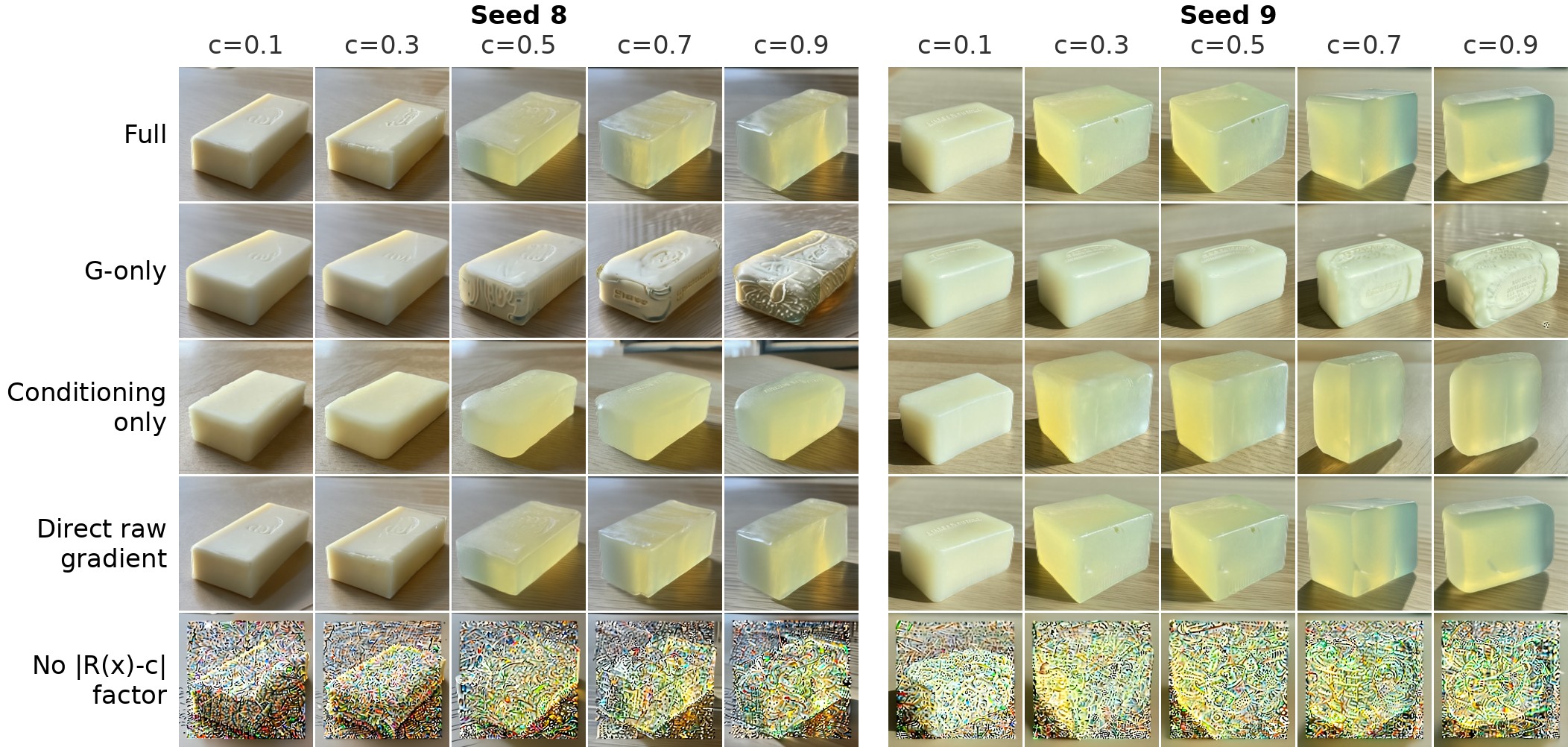}{Soap translucency, ablation, test seeds 8--9.}{fig:abl-soap-3}


\small
\begin{thebibliography}{99}
\bibitem{Bansal2024} A. Bansal et al. Universal guidance for diffusion models. In \emph{ICLR}, 2024.
\bibitem{Baumann2025} S. A. Baumann et al. Continuous, subject-specific attribute control in T2I models by identifying semantic directions. In \emph{CVPR}, pages 13231--13241, 2025.
\bibitem{Bloch1995} P. H. Bloch. Seeking the ideal form: Product design and consumer response. \emph{Journal of Marketing}, 59(3):16--29, 1995.
\bibitem{Burnap2023} A. Burnap, J. R. Hauser, and A. Timoshenko. Product aesthetic design: A machine learning augmentation. \emph{Marketing Science}, 42(6):1029--1056, 2023.
\bibitem{Cawley2010} G. C. Cawley and N. L. C. Talbot. On over-fitting in model selection and subsequent selection bias in performance evaluation. \emph{Journal of Machine Learning Research}, 11:2079--2107, 2010.
\bibitem{Chen2026} D. Chen et al. AttriCtrl: A generalizable framework for controlling semantic attribute intensity in diffusion models. In \emph{ICLR}, 2026.
\bibitem{Chiu2026} P.-Y. Chiu, I.-S. Fang, and J.-C. Chen. Text Slider: Efficient and plug-and-play continuous concept control for image/video synthesis via LoRA adapters. In \emph{WACV}, pages 613--622, 2026.
\bibitem{Cho2025} W. Cho et al. Att-Adapter: A robust and precise domain-specific multi-attributes T2I diffusion adapter via conditional VAE. In \emph{ICCV}, pages 15626--15635, 2025.
\bibitem{Chung2023} H. Chung et al. Diffusion posterior sampling for general noisy inverse problems. In \emph{ICLR}, 2023.
\bibitem{Clark2024} K. Clark et al. Directly fine-tuning diffusion models on differentiable rewards. In \emph{ICLR}, 2024.
\bibitem{Crilly2004} N. Crilly, J. Moultrie, and P. J. Clarkson. Seeing things: Consumer response to the visual domain in product design. \emph{Design Studies}, 25(6):547--577, 2004.
\bibitem{Dhariwal2021} P. Dhariwal and A. Nichol. Diffusion models beat GANs on image synthesis. In \emph{NeurIPS}, 2021.
\bibitem{Ding2021} X. Ding et al. CcGAN: Continuous conditional generative adversarial networks for image generation. In \emph{ICLR}, 2021.
\bibitem{Ding2023} X. Ding et al. Continuous conditional generative adversarial networks: Novel empirical losses and label input mechanisms. \emph{IEEE Transactions on Pattern Analysis and Machine Intelligence}, 45(7):8143--8158, 2023.
\bibitem{Ding2024} X. Ding et al. CCDM: Continuous conditional diffusion models for image generation. \emph{IEEE Transactions on Multimedia}, 28:4360--4372, 2026.
\bibitem{Ding2026} X. Ding et al. Enhancing diffusion-based quantitatively controllable image generation via matrix-form EDM and adaptive vicinal training. arXiv:2602.02114, 2026.
\bibitem{Esser2024} P. Esser et al. Scaling rectified flow transformers for high-resolution image synthesis. In \emph{ICML}, 2024.
\bibitem{Ezra2025} R. Ezra et al. FreeSliders: Training-free, modality-agnostic concept sliders for fine-grained diffusion control in images, audio, and video. arXiv:2511.00103, 2025.
\bibitem{Gandikota2024} R. Gandikota et al. Concept sliders: LoRA adaptors for precise control in diffusion models. In \emph{ECCV}, pages 172--188, 2024.
\bibitem{Gao2023} L. Gao, J. Schulman, and J. Hilton. Scaling laws for reward model overoptimization. In \emph{ICML}, pages 10835--10866, 2023.
\bibitem{Gebhardt2025} C. Gebhardt et al. Regressor-guided generative image editing balances user emotions to reduce time spent online. arXiv:2501.12289, 2025.
\bibitem{Hayakawa2024} R. Hayakawa, R. Noji, and T. Kato. Generating tyre tread designs using a sensory evaluation regression model and a generative model. \emph{Journal of Engineering Design}, 36(1):19--51, 2025.
\bibitem{He2024} Y. He et al. Manifold preserving guided diffusion. In \emph{ICLR}, 2024.
\bibitem{Ho2022} J. Ho and T. Salimans. Classifier-free diffusion guidance. arXiv:2207.12598, 2022.
\bibitem{Hu2022} E. J. Hu et al. LoRA: Low-rank adaptation of large language models. In \emph{ICLR}, 2022.
\bibitem{Hu2023} Y. Hu, B. Liu, J. Kasai, Y. Wang, M. Ostendorf, R. Krishna, and N. A. Smith. TIFA: Accurate and interpretable text-to-image faithfulness evaluation with question answering. In \emph{ICCV}, pages 20406--20417, 2023.
\bibitem{Ibarrola2024} F. Ibarrola, R. Lulham, and K. Grace. Affect-conditioned image generation. \emph{IEEE Transactions on Affective Computing}, 15(4):2169--2179, 2024.
\bibitem{Liao2023} C. Liao, M. Sawayama, and B. Xiao. Unsupervised learning reveals interpretable latent representations for translucency perception. \emph{PLoS Computational Biology}, 19(2):e1010878, 2023.
\bibitem{Oquab2023} M. Oquab et al. DINOv2: Learning robust visual features without supervision. \emph{Transactions on Machine Learning Research}, 2024.
\bibitem{Pizzi2022} E. Pizzi, S. D. Roy, S. N. Ravindra, P. Goyal, and M. Douze. A self-supervised descriptor for image copy detection. In \emph{CVPR}, pages 14512--14522, 2022.
\bibitem{Radford2021} A. Radford et al. Learning transferable visual models from natural language supervision. In \emph{ICML}, 2021.
\bibitem{Saraee2018} E. Saraee, M. Jalal, and M. Betke. SAVOIAS: A diverse, multi-category visual complexity dataset. arXiv:1810.01771, 2018.
\bibitem{Shen2024} Y. Shen et al. Understanding and improving training-free loss-based diffusion guidance. In \emph{NeurIPS}, 2024.
\bibitem{Simeoni2025} O. Sim\'eoni et al. DINOv3. arXiv:2508.10104, 2025.
\bibitem{Song2023} J. Song et al. Loss-guided diffusion models for plug-and-play controllable generation. In \emph{ICML}, 2023.
\bibitem{Toet2019} A. Toet et al. CROCUFID: A cross-cultural food image database for research on food elicited affective responses. \emph{Frontiers in Psychology}, 10:58, 2019.
\bibitem{Yang2023} C. Yang, F. Liu, and J. Ye. A product form design method integrating Kansei engineering and diffusion model. \emph{Advanced Engineering Informatics}, 57:102058, 2023.
\bibitem{Ye2024} H. Ye et al. TFG: Unified training-free guidance for diffusion models. In \emph{NeurIPS}, 2024.
\bibitem{Ye2026} W. Ye et al. All-in-One Slider for attribute manipulation in diffusion models. In \emph{CVPR}, pages 18693--18702, 2026.
\bibitem{Zarei2026} A. Zarei et al. SliderEdit: Continuous image editing with fine-grained instruction control. In \emph{CVPR}, pages 14430--14439, 2026.
\bibitem{Zhai2023} X. Zhai et al. Sigmoid loss for language image pre-training. In \emph{ICCV}, pages 11975--11986, 2023.
\bibitem{Zhang2024} Z. Zhang et al. Confronting reward overoptimization for diffusion models: A perspective of inductive and primacy biases. In \emph{ICML}, pages 60396--60413, 2024.
\bibitem{Zhu2025} Z. Zhu et al. CompSlider: Compositional slider for disentangled multiple-attribute image generation. In \emph{ICCV}, pages 16682--16691, 2025.
\end{thebibliography}
\end{document}